%% file: compspace_arXiv.tex
\documentclass[12pt]{article} 
\usepackage{amsthm}
\usepackage{amsmath,amssymb}      
\usepackage{skak}
\usepackage{chessboard} 
\usepackage{stmaryrd}
\usepackage[normalem]{ulem}

\usepackage{xspace}
\usepackage{color}
\usepackage{epsfig}     
\graphicspath{{.}{./figures/}}  

\usepackage{macros/tikzfig}
\usepackage{keycommand}

\input{macros/defs.tex}

\input{macros/tikzstyles.tex}

\input{macros/tikzfigures.tex}
\definecolor{hexcolor0xa9a9a9}{rgb}{0.663,0.663,0.663} 
\tikzstyle{GrayLine}=[dashed,draw=hexcolor0xa9a9a9] 
\tikzstyle{gray}=[dashed,draw=hexcolor0xa9a9a9]

\theoremstyle{definition}
\newtheorem{theorem}{Theorem}[section]
\newtheorem*{theorem*}{Theorem}

\newtheorem{example*}[theorem]{Example*}
\newtheorem{examples*}[theorem]{Examples*}
\newtheorem{remark}[theorem]{Remark}
\newtheorem{remark*}[theorem]{Remark*}

\def\bR{\begin{color}{red}}  
\def\bB{\begin{color}{blue}}
\def\bM{\begin{color}{magenta}}  
\def\bC{\begin{color}{cyan}}
\def\bW{\begin{color}{white}}
\def\bBl{\begin{color}{black}}
\def\bG{\begin{color}{green}}
\def\bY{\begin{color}{yellow}}
\def\e{\end{color}\xspace}
\newcommand{\bit}{\begin{itemize}}
\newcommand{\eit}{\end{itemize}\par\noindent}
\newcommand{\ben}{\begin{enumerate}}
\newcommand{\een}{\end{enumerate}\par\noindent}
\newcommand{\beq}{\begin{equation}}
\newcommand{\eeq}{\end{equation}\par\noindent}
\newcommand{\beqa}{\begin{eqnarray*}}
\newcommand{\eeqa}{\end{eqnarray*}\par\noindent}
\newcommand{\beqn}{\begin{eqnarray}}
\newcommand{\eeqn}{\end{eqnarray}\par\noindent}

\title{Talking Space:\\
inference from spatial linguistic  meanings}

\author{Vincent Wang-Ma\'{s}cianica${}^{\dagger\ddagger}$ and Bob Coecke${}^{\dagger}$ \\  
${}^{\dagger}$Cambridge Quantum Computing Ltd.\\ 
${}^{\ddagger}$Oxford University, Department of Computer Science}
 
\begin{document}
\maketitle

\begin{abstract}   
This paper concerns the intersection of natural language and the physical space around us in which we live, that we observe and/or imagine things within.  Many important features of language  have  spatial connotations, for example, many prepositions (like in, next to, after, on, etc.) are fundamentally spatial. Space is also a key factor of the meanings of many words/phrases/sentences/text, and space is a, if not the key, context for referencing  (e.g.~pointing) and embodiment. 

We propose a mechanism for how space and  linguistic structure can be made to interact in a matching compositional fashion. Examples include Cartesian space, subway stations, chesspieces on a chess-board, and Penrose's staircase. The starting point for our construction is the DisCoCat model of compositional natural language meaning, which we relax to accommodate physical space. We address the issue of having multiple agents/objects in a space, including the case that each agent has different capabilities with respect to that space, e.g.,~the specific moves each chesspiece can make, or the different velocities one may be able to reach.
 
Once our model is in place, we show how inferences drawing from the structure of physical space can be made. We also how how linguistic model of space can interact with other such models related to our senses and/or embodiment, such as the conceptual spaces of colour, taste and smell, resulting in a rich compositional model of meaning that is close to human experience and embodiment in the world. 
\end{abstract}
 
\section{Introduction}

The physical space around us -- that is, the space which we observe around us, in which we interact with others  and move within -- is arguably the most important context for human experience. It should come as no surprise that it plays a central role in the use and development of language \cite{lakoff_metaphors_2003}. Moreover, as G\"ardenfors \cite{gardenfors2, gardenfors} points out, physical space is a natural mediator between sensory data and linguistic content: for many of us, interpreting linguistic content is spatial visualisation.
 
A naive perception of space could be a base plane with an upward direction:
\beq\label{humanspace} 
\input{space.tikz} 
\eeq
referring to where we move around, as well as an upward direction when looking at the birds, and a downward direction whenever we start digging holes.  

Mathematically speaking, this na\"{i}ve perceptive space  -- a more detailed treatment of which can be found in \cite{zwarts_locative_2016} --  can easily be embedded within three-dimensional Cartesian space $\mathbb{R}^3$. There are some alternative options of what to mean by space.  If you happen to be a chesspiece on a chess board, then of course the space that you experience is an 8 by 8 grid, and if you are taking a subway the relevant notion of space is the different stations along that line.  But all the essential notions here can be embedded in $\mathbb{R}^3$, the chess board being a restriction and discretisation of $\mathbb{R}^2$, and the subway line being the same for $\mathbb{R}^1$, after yanking the track straight. 

In making space explicit in a compositional theory of language meaning, one expects to make inferences that require `awareness' of the structure of space. What we mean by this is best explained by examples. If we say that:
\bit
\item {\tt The painting is above the chest}
\item {\tt The light is above the painting}
\eit 
then one wants to be able to infer that: 
\bit
\item {\tt The light is above the chest}
\eit
Similarly, if:
\bit
\item {\tt Alice chases Bob}
\item {\tt Alice is in Paris} 
\eit
then one want to be able to infer that: 
\bit
\item {\tt Bob is in Paris}
\eit

In order to combine physical space and linguistic structure we will take the DisCoCat model of compositional natural language meaning \cite{CSC} as our starting point. In its most general interpretation, it assumes a compositional model of linguistic structure, and meaning spaces organised in a matching structure. 
 
For the purposes of this paper, we can do with a representation of grammar in terms of pregroups \cite{Lambek1, LambekBook}, a simple syntactic formalism with an appealing diagrammatic presentation. As shown in \cite{DRichie}, also a more expressive formalism such as Combinatory Categorial Grammar \cite{steedman1987combinatory} can be equipped with a mapping to a semantic category of the type that we assume here. Other linguistic structure indentified specifically within the DisCoCat approach, most notably `internal wirings' for relative pronouns \cite{FrobMeanI, FrobMeanII}, verbs \cite{GrefSadr, KartsaklisSadrzadeh2014, CoeckeText, CoeckeMeich} and other grammatical types \cite{GramEqs, GramCircs}, will also have counterparts in our compositional model of space.

The central notion of our model will be spatial relations, which express how objects/agents are related within  that space.  In order to compose these  spatial relations  across different sentences, it will be necessary to pass from DisCoCat to DisCoCirc \cite{CoeckeText, GramCircs},  which deals with the composition of sentences, and how that composition updates the meanings of multiple objects/agents. We also address the modelling of spatial features that are specific to objects/agents, such as the moves available to specific chesspieces or the spatial extent of an object, and how they can be encoded within that agent.

We then move on to explain how  a language diagram like:  
\[
\input{relpronchessd.tikz} 
\]
can specify a specific piece on a chessboard, exposing explicit `spatial reasoning capability' of our model.

\paragraph{Related work.} In appendix \ref{app} we outline the relationship to Montague semantics, not just of this paper, but the DisCoCat program more broadly.

We are \underline{not} proposing a rule-based account of spatial relations, which has been done many times before \cite{vieu_spatial_1997,casati_spatial_1997}, e.g.~systematically enforcing that \texttt{above} must be transitive.   These relations are built-into our model, as we will demonstrate.   Such an approach is better aligned with modern AI practices than rule-based systems are.

We are \underline{not} proposing mechanisms that bridge linguistic and spatial-cognitive faculties, which is a deeply-studied field in its own right \cite{herskovits_language_1997}. So, we don't address questions of cognitive compliance.

We \underline{do} propose, to the best of our knowledge, the first linguistic mathematical formalism of space: which is to say that the compositional structure of our spatial content matches the compositional structure of spatially descriptive natural language.  This level of compositional concern lets us remain ``implementation agnostic" at the level of individual spatial relations, which can be learnt by statistical methods \cite{dawson_generative_2013, collell_learning_2018}, or specified by-hand, as we do for illustrative examples to follow.

%
%
%


\section{Spatial relations}

We provide some examples of space that we will consider throughout this paper, how one can understand relations `within' these different spaces, and essential operations upon relations that we will require later.

\subsection{Examples of spaces}

\paragraph{Cartesian space.} Cartesian space is the familiar 3-dimensional space we live in. We typically write Cartesian space as $\mathbb{R}^3$, to refer to points as represented by coordinates $(x,y,z)$ for $x, y, z\in\mathbb{R}$, corresponding to directions in Cartesian space, typically backward-forward, left-right, and down-up, respectively. Doing so requires one to pick an origin $(0,0,0)$. So, the coordinate $(-4,2,0.69)$ can be interpreted as movement instructions from the origin to get to the point that the coordinate names: ``starting from the origin, move 4 units backward, 2 units right, and 0.69 units upward.''

\paragraph{Cartesian space-time.} This is very much similar to the previous example, but now with one additional dimension, namely time.  We can write this as $\mathbb{R}^4$ or as $\mathbb{R}^3\times\mathbb{R}$ to stress that time plays a different role. Now the coordinates are 4-tuples $(x,y,x,t)$.

\paragraph{Subway line.} For the Hong Kong subway system MTR, just as for any other subway system, a colour represents a subway line and the dots on the line refer to stations: 
\[
\epsfig{figure=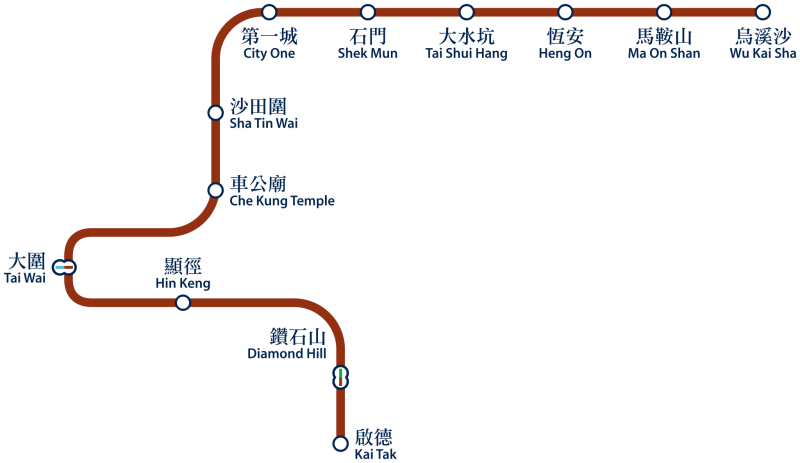,width=240pt}    
\]
The `brown' Tuen Ma line, has the following set of stations:
\[
\texttt{Tuen Ma} := 
\{ \texttt{Kai Tak} \ , \ \texttt{D.~H.} \ , \ \texttt{Hin Keng} \ , \ldots, \ \texttt{Ma On Shan} \ , \ \texttt{Wu Kai Sha} \}
\]

This space is one-dimensional and finite, hence discrete.  The distances between different stations differ, which defines the structure of the space, and this structure can still be captured by embedding it in Cartesian space.  However, it may make more sense to conceive of it as a space in its own right, and specify its structure as an ordering of these stations (i.e.~``\texttt{Ma On Shan} is $n$ stations away from here''), possibly also giving the travel times between stations  (i.e.~``\texttt{Ma On Shan} is $t$ minutes away from here'').
 
\paragraph{Chessboard.} A chessboard is a square subdivided on each side into 8 equal lengths:
\[
\scalebox{0.75}{
\chessboard[pgfstyle=straightmove,shortenstart=1ex,arrow=to,linewidth=0.1ex,color=red,shortenstart=1ex,showmover=false]
}
\]
to give $8 \times 8 = 64$ squares. We can denote the set of all these squares as follows:
\[\texttt{chessboard} := \{\texttt{a1} \ , \ \texttt{a2} , \ \ldots \ , \ \texttt{a8} \ , \ \texttt{b1} \ , \ \texttt{b2} , \ \ldots \ , \ \texttt{b8} \ , \ldots \ , \ \texttt{h1} \ , \ \texttt{h2} , \ \ldots \ , \ \texttt{h8}\}
\]
We can think of this as the Cartesian product: 
\[
X \times Y := \{(x,y) \ | \ x \in X, y \in Y \}
\]
of two sets $\texttt{a-h} := \{ \texttt{a} \ , \ \texttt{b} \ , \ \ldots\ , \texttt{h} \}$ and $\texttt{1-8} := \{ \texttt{1} , \texttt{2}, \ldots, \texttt{8}\}$ so that:
\[
\texttt{chessboard} := \texttt{a-h} \times \texttt{1-8}
\]

This space can be embedded in Cartesian space, and one can think of it as a two-dimensional analog to the subway line, if all we care about in the latter case is the ordering of the stations. What makes this space special are its inhabitants. Each of the chesspieces can only make a particular set of moves, for example, for a {\tt knight} the nearest neighbours of its locations are not the squares it can most easily get to:
\[
\scalebox{0.75}{
\chessboard[pgfstyle=4x4,setwhite={Nd4},pgfstyle=knightmove,arrow=to,linewidth=0.1ex,color=red,pgfstyle=knightmove,markmoves={d4-e6,d4-f3,d4-b3,d4-c2,d4-c6,d4-b5,d4-e2,d4-f5},shortenstart=1ex,showmover=false]
}
\]

\paragraph{Penrose's staircase.} This is an example of something that can no longer be embedded in Cartesian space, but our general model can still handle it.
\[
\epsfig{figure=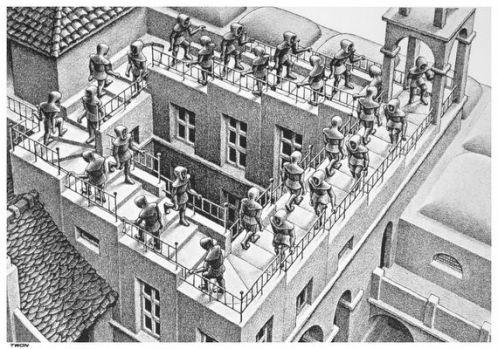,width=200pt}   
\]
We represent the staircase as a set as four staircases $\texttt{I-IV} := \{ \texttt{I} \ , \ \texttt{II} \ , \ \texttt{III} \ ,\ \texttt{IV} \}$, each with steps $\texttt{1-n} := \{ \texttt{1} , \texttt{2}, \ldots, \texttt{n}\}$, not accounting the lowest step for each of these:
\[
\texttt{Penrose} := \texttt{I-IV} \times \texttt{1-n} 
\]
 
\subsection{Examples of spatial relations}\label{sec:spacialrelations}

Relation(-ship) is a very intuitive notion, that allows one to express many important concepts, for example, here in the case of {\tt The Simpsons}:
\[ 
\texttt{siblings} := \{ (\texttt{Bart}, \texttt{Lisa}) \ , \ (\texttt{Lisa},\texttt{Bart}) \ , \ (\texttt{Lisa},\texttt{Maggie}) \ , \ \ldots \}  
\]
Relations are the obvious concept for discussing spatial features related to the `inhabitants' of that space. Formally, a relation between a set $X$ and itself takes the form of a subset of the Cartesian product: 
\[
R \subseteq X \times X
\]
that is, it is a set of pairs $(x, y)$ of elements $x, y\in X$.  We can specify that a pair is in the relation by writing $R(x, y)$. We can generalise this to triples: 
\[
R \subseteq X \times X\times X
\]
or arbitrary $n$-tuples, and also to single spaces: 
\[
R \subseteq X  
\]
which gives us plain subsets of $X$.  Rather than describing `inter'-relationships between two or more parties, these `unary' relations let us say something about a particular inhabitant itself, e.g.~where that inhabitant is. We now show how we can assign meaning to these relations `within' our example spaces. 

\paragraph{Relations for a subway line.} Here an obvious example is:  
\[
\texttt{next stop}  := \{ (\texttt{Kai Tak},\texttt{D.~H.}) \ , \ (\texttt{D.~H.},\texttt{Hin Keng}) \ , \ldots, \ (\texttt{Ma on Shan},\texttt{Wu Kai Sha}) \}
\]
Here is one that relates 3 stations:
\[
\texttt{in-between} :=\{ (\texttt{Kai Tak},\texttt{D.~H.}, \texttt{Hin Keng}) \ , \ (\texttt{Kai Tak}, \texttt{Ma On Shan}, \texttt{Wu Kai Sha}) \ , \ldots \} 
\]
and here is one about 1 station:
\[
\texttt{my\ station} :=\{\texttt{Wu Kai Sha}\} 
\]

\paragraph{Relations for a chessboard.} Here's a graphical representation of a relation that is an analogue to one that we saw above:
\[
\texttt{move\ right} :=
\scalebox{0.75}{\raisebox{-2.7cm}{
\chessboard[pgfstyle=straightmove,shortenstart=0.5ex,arrow=to,linewidth=0.1ex,color=red,markmoves={a1-b1,a2-b2,a3-b3,a4-b4,a5-b5,a6-b6,a7-b7,a8-b8,b1-c1,b2-c2,b3-c3,b4-c4,b5-c5,b6-c6,b7-c7,b8-c8,c1-d1,c2-d2,c3-d3,c4-d4,c5-d5,c6-d6,c7-d7,c8-d8,d1-e1,d2-e2,d3-e3,d4-e4,d5-e5,d6-e6,d7-e7,d8-e8,e1-f1,e2-f2,e3-f3,e4-f4,e5-f5,e6-f6,e7-f7,e8-f8,f1-g1,f2-g2,f3-g3,f4-g4,f5-g5,f6-g6,f7-g7,f8-g8,g1-h1,g2-h2,g3-h3,g4-h4,g5-h5,g6-h6,g7-h7,g8-h8},shortenstart=1ex,showmover=false]
}
}
\]
There also are the chesspiece specific moves, for example for a king: 
\beq\label{kingsmoves}
\texttt{king's moves} :=
\scalebox{0.75}{\raisebox{-2.7cm}{
\chessboard[pgfstyle=straightmove,shortenstart=0.5ex,arrow=to,linewidth=0.1ex,color=red,markmoves={e1-d1,e1-d2,e1-e2,e1-f2,e1-f1,c5-c4,c5-c6,c5-b4,c5-b5,c5-b6,c5-d4,c5-d5,c5-d6,h8-g8,h8-g7,h8-h7},shortenstart=1ex,showmover=false,setwhite={Ke1,Kc5,Kh8}]
}
}
\eeq
which concretely would be the relation made up of all pairs of adjacent squares.  Note that {\tt next to} corresponds to the same relation. We can also symbolically formalise this:
\beq\label{eq:nextto}
{\tt next\ to} := \left\{(\alpha x, \beta y) \Bigm| 
\begin{array}{c}
\alpha, \beta \in \texttt{a-h}\\
x, y\in \texttt{1-8} \end{array}, 
\begin{array}{r}   
|\alpha - \beta| \times |x - y| \leq 1\\ 
\mbox{ not } |\alpha - \beta| = |x - y| = 0 
\end{array}
\right\}
\eeq

\paragraph{Relations for Penrose's staircase.} These are pretty much all there is:
\[
{\tt move\ 1\ step\ up} := \{({\tt I1}, {\tt I2}), \ldots, ({\tt I(n-1)}, {\tt In}), ({\tt In}, {\tt II1}), \ldots, ({\tt IVn}, {\tt I1}) \}
\]
\[
{\tt move\ 1\ step\ down} := \{({\tt In}, {\tt I(n-1)}), \ldots, ({\tt I2}, {\tt I1}), ({\tt I1}, {\tt IVn}), \ldots, ({\tt II1}, {\tt In}) \}
\]

\paragraph{Relations for Cartesian space.} These are intuitive because we live here, alongside many other shapes and forms. However, to keep things simple in terms of mathematical description, in many cases it is convenient to think of all of the inhabitants of our space as a points, just like physicists always do. We can specify directions and distances between two points. Here are some examples of prepositions:
\[
{\tt higher\ than} := \Bigl\{\bigl((x,y,z),(x',y',z') \bigr)\Bigm| z > z'   \Bigr\} 
\] 
\beq\label{above}
{\tt above} := \Bigl\{\bigl((x,y,z),(x',y',z') \bigr)\Bigm|x=x'\ ,\ y=y'\ ,\  z > z'   \Bigr\}
\eeq
Given that we essentially move on the Earth's surface:
\[
{\tt close\ to} := \Bigl\{\bigl((x,y,z),(x',y',z') \bigr)\Bigm||(x,y),(x',y')|\leq\epsilon\ ,\  z = z'   \Bigr\}
\] 
Again we have this one involving three parties:
\[
\texttt{in-between} := \Bigl\{\bigl((x,y,z),(x',y',z'),(x'',y'',z'') \bigr)\Bigm| 
\exists p\in (0,1) \mbox{ such that } (*)  \Bigr\} 
\] 
where $(*)$ means $(x',y',z')= p\cdot(x,y,z) + (1-p)\cdot(x'',y'',z'')$.

If ${\cal R}\subset X$ is a  region in space then we get an adjective:
\[
{\tt in\ {\cal R}} := \Bigl\{(x,y,z) \Bigm| (x,y,z)\in  {\cal R} \Bigr\}  = {\cal R}
\] 
and here is an example of a family of spatio-temporal verbs. We omit the subscript where it is clear from context, or where it doesn't matter:
\beq\label{ChaseCart}
{\tt chases}_{\delta t > 0} := \Bigl\{\bigl((x,y,z, t),(x,y,z, t') \bigr)\Bigm| t = t' + \delta t   \Bigr\} 
\eeq

\subsection{Graphical presentation of relations} 

We now provide a standard graphical presentation of relations.  This will become particularly handy in the next section, when we will start composing relations.  The graphical representation also elucidates a fundamental structure that governs relations.

We represent a relation of type $R\subseteq X \times X$ as a \emph{box}:
\ctikzfig{boxsingle}
We think of the \emph{wires} as representing the set $X$.  As we shall see below, there can also be more general boxes with multiple inputs and outputs: 
\ctikzfig{box}
and there are two special cases when there either are no inputs or no outputs:  
\ctikzfig{stateeffect}
called \emph{state} and \emph{test} respectively.      When we plug such boxes together:
\ctikzfig{boxexample}
we get a \emph{diagram} and will again explain later what that plugging means in terms of relations.  Importantly, we read these diagrams in this paper from top to bottom.

We will label the box with the relation it represents e.g.:
\ctikzfig{boxR}
for $R\subseteq X \times X$, where we think of the first $X$ in $X \times X$ as the input and the second $X$ as the output.  For a subset $Q\subseteq X$ there only is one $X$ so we can represent it as a state:
\ctikzfig{stateRsingle}

But of course, we can also think of $R\subseteq X \times X$ as a subset of $X \times X$, and the same goes for $S\subseteq X \times X \times X$,  so we could also represent these as:
\ctikzfig{stateR}
At the same time, we could think of $S\subseteq X \times X \times X$ as a proper box in more than one manner, e.g.~$S\subseteq (X \times X) \times X$ and $S\subseteq X \times (X \times X)$:
\ctikzfig{stateRbis}
This fact, that we can think of relations in different, equivalent ways is a key feature of relations. It is of course also a burden as it causes ambiguity of what we are actually talking about when we denote relations as $R\subseteq X \times \ldots \times X$.  

This ambiguity goes away when we draw the diagrams, and we can eliminate it symbolically, as we already did above, by using brackets.  To distinguish between states and tests, 
we introduce the singleton set $\{*\}$ so that we have:
\[
{\tt state}\subseteq \{*\}\times (X \times X)\qquad\qquad\qquad\qquad {\tt test}\subseteq  (X \times X)\times \{*\}
\]
which is justified by these isomorphisms:
\[
\{*\}\times (X \times X)\ \simeq\  X\times X\  \simeq\  (X \times X)\times \{*\}
\] 

Once disambiguated, we can  turn the different representations of a relation into each other using two very special relations, a \emph{cap-state} and an \emph{cup-test} respectively:
\[
\input{cap.tikz}\ :=\Bigl\{\bigl((*,(x,x)\bigr) \Bigm| x\in X\Bigr\}\qquad\qquad\qquad
\input{cup.tikz}\ :=\Bigl\{\bigl((x,x),*\bigr) \Bigm| x\in X\Bigr\} 
\]
We will show below how exactly that conversion works.   More importantly, these two special relations will  be key to expressing grammatical structure as relations.  The reason for their notations is that they obey a certain equality that shows that they effectively `behave' like bent wires, as we will show later. 

These special relations are instances of a larger family of special relations called \emph{spiders} \cite{CKbook}.  They are defined as follows:
\[
\input{spidercomp.tikz}\ :=\ \Bigl\{\bigl((x, \ldots, x),(x, \ldots, x) \bigr)\Bigm|x\in X   \Bigr\}
\]
and their key property is:
\beq\label{eq:fusion}
\input{spider.tikz} \ = \ \input{spidercomp.tikz}   
\eeq
One also easily checks the following:
\beq\label{eq:fusion2}
\input{plaindot.tikz}\ \ =\ \ \input{plain.tikz}\qquad\qquad\quad\input{cupdot.tikz} \ \ =\ \ \input{cup.tikz}\qquad\qquad\quad\input{capdot.tikz} \ \ =\ \ \input{cap.tikz}
\eeq
and these special cases \emph{copy} and \emph{delete}:
\[
\input{copy.tikz}\ =\ \Bigl\{\bigl(x,(x, x) \bigr)\Bigm|x\in X   \Bigr\}\qquad\quad\qquad
\input{delete.tikz}\ =\ \Bigl\{\bigl(x,* \bigr)\Bigm|x\in X   \Bigr\} 
\]
also give an intuitive understanding.  Furthermore: 
\beq\label{intersection}
\input{AND1.tikz}\ =\ \input{AND2.tikz}
\eeq
and:
\beq\label{unknown}
\input{unknown.tikz}\ =\ \Bigl\{\bigl(*,x \bigr)\Bigm|x\in X   \Bigr\}\ \simeq X
\eeq
so we can call these \emph{AND} and \emph{unknown} respectively.

\subsection{Composing (spatial) relations} 

We can build new spatial relations from old ones, using the standard operations of relations, sequential composition and parallel composition. An intuitive mathematical presentation of these operations can be found in \cite{CatsII, CKbook}.   

\paragraph{Basic sequential composition (a.k.a.~`after').} For relations of type $R, S \subseteq X \times X$ sequential composition  is defined as follows:
\beq\label{seqcomp}
S\circ R:= \Bigl\{(x,x'') \Bigm| \exists x' \mbox{ such that } R(x, x') \mbox{ and } S(x', x'')\Bigr\} 
\eeq
In terms of Venn-diagrams this looks as follows:
\[
\input{compose_venn.tikz}
\]
that is, we have a path from $x$ to $x''$ (i.e.~$(S\circ R)(x, x'')$) whenever there exists a path from $x$ to $x''$, passing through some $x'$ (i.e.~$R(x, x')$ and $S(x', x'')$).  In diagrams, we denote it by connecting the corresponding boxes:
\[
\input{comp.tikz}
\]

As a first example, doing \texttt{next stop} twice gets us two stops further: 
\[
\texttt{next stop}\circ \texttt{next stop}
= \{ (\texttt{Kai Tak},\texttt{Hin Keng}) \ ,  \ldots, \ (\texttt{Heng On},\texttt{Wu Kai Sha}) \}
\]
we can do this many times, that is $\texttt{next stop}^n$, but as there only are 12 stops in total, there in no way to get 12 steps further:
\[
(\texttt{next stop})^{12}= \{  \}
\]
On the other hand in the case of Penrose's staircase we can keep going upstairs forever, and in particular do we have:
\[
(\texttt{\tt move\ 1\ step\ up})^{4n} = \{({\tt I1}, {\tt I1}), ({\tt I2}, {\tt I2}), \ldots, ({\tt IVn}, {\tt IVn}) \}
\] 
Similar examples exist in Cartesian space.

\paragraph{More general sequential composition.} The above defined composition is only one instance of sequential composition for relations.  One important example is \emph{application} of a relation $S \subseteq X \times X$ to a subset $R\subseteq X$, defined as follows: 
\beqn\label{eq:application}
\input{comp2.tikz}\ &:=& \Bigl\{(*,x') \Bigm| \exists x \mbox{ such that } R(*, x) \mbox{ and } S(x, x')\Bigr\}\\
             &\simeq& \Bigl\{x' \Bigm| x'\in R \mbox{ and } S(x, x')\Bigr\}
\eeqn
where the first line gives the analogue to (\ref{seqcomp}) using the singleton-set in the state representation, while the second line has states as plain subsets.

In the case of the chessboard example, for $R=\{{\tt c3}\}$ we have:
\[
\texttt{king's moves}\,\circ R =  
\scalebox{0.75}{
\raisebox{-2.7cm}{\chessboard[pgfstyle=straightmove,shortenstart=1ex,arrow=to,linewidth=0.1ex,color=red,shortenstart=1ex,showmover=false, color=green!50, colorbackfields={b2, c2, d2, b3, d3, b4, c4, d4}]} }
\]
that is, all the places the king can end up from {\tt c3} after one move.  

In Cartesian space we have for $R:=\{(0,0,0)\}$:
\[
\texttt{higher\ than}\,\circ R = \Bigl\{(x',y',z') \Bigm| z' > 0   \Bigr\}
\] 
which gives those points higher than the `origin' $R$.  

\paragraph{Multiple inhabitants.} Parallel composition operations require multiple spaces in parallel, which here means a Cartesian product $X\times \ldots \times X$ of those spaces.  As we are focusing on inhabitants of a particular space, for each of these inhabitants, we need one copy of that space to represent that inhabitant. This bears repeating:
\[
\boxed{\text{We assign each inhabitant their own copy of the space.}}
\]
In fact, we already implicitly have been using this in the examples of Section \ref{sec:spacialrelations}.  Moreover, as we shall see later, grammatical structure also requires multiple copies of the spaces.

One way to specify which inhabitant we are talking about is to tag everything related to `their' space with that inhabitant's name. For example, if {\tt Alice} and {\tt Bob} both take the subway, then the space we need to consider is: 
\[
\texttt{Tuen Ma}_{\tt Alice} \times \texttt{Tuen Ma}_{\tt Bob} 
\]
For {\tt Alice} and {\tt Bob} just hanging around in Cartesian space, then we consider: 
\[
\mathbb{R}^3_{\tt Alice} \times \mathbb{R}^3_{\tt Bob} 
\]
This larger space increases expressiveness as we can now have relations involving multiple inhabitants. For example if {\tt Alice} and {\tt Bob} travel together on \texttt{Tuen Ma}:
\beqa
\texttt{Alice with Bob}    
&:=& \Bigl\{ \bigl(*,(\texttt{Kai Tak},\texttt{Kai Tak})\bigr) \ , \ldots, \ \bigl(*,(\texttt{Wu Kai Sha},\texttt{Wu Kai Sha})\bigr) \Bigr\}\\
&:=& \{ *\}\times \Bigl\{ (\texttt{Kai Tak},\texttt{Kai Tak}) \ , \ldots, \ (\texttt{Wu Kai Sha},\texttt{Wu Kai Sha}) \Bigr\}\\
&\ \subseteq& \{ *\}\times \left(\texttt{Tuen Ma}_{\tt Alice} \times \texttt{Tuen Ma}_{\tt Bob}\right)
\eeqa
where we used $\{*\}$ to indicate that we are describing a state:
\[
\input{stateAB.tikz}
\]
and not a regular box. Our chessboard can now also have more than one piece:
\[
\scalebox{0.75}{
\chessboard[pgfstyle=straightmove,shortenstart=1ex,arrow=to,linewidth=0.1ex,color=red,shortenstart=1ex,showmover=false,setfen=4r3/2n2k2/P3p1p1/5p2/1P1K3N/2PQ4/r4B2/8 w - - 0 20,
            pgfstyle=border]
}
\] 

For many purposes this labelling will be sufficient. However, in the case of a chessboard,  the label of chesspieces doesn't tell us all the relevant spatial information of that inhabitant. This is important since different pieces can make different moves. Similarly, in Cartesian space, some can run faster than others, and when we go beyond representing things as points, sizes are different.  We address this in Section (\ref{varyinginhab}).

\paragraph{Parallel composition or (a.k.a.~`while').} In terms of diagrams this is: 
\[
\input{parallel.tikz}
\]
In concrete terms, for relations $R, S\subseteq X\times X$ the result will be a relation $R\times S:  (X\times X)\times(X\times X)$, where the bracketing here is about the parallel aspect, and not the different ways of representing a relation discussed above.  Diagrams perfectly disambiguate the confusion that may arise here. Concretely, we have:
\beq\label{seqcomp}
S\times R:= \Bigl\{\bigl((x,x'),(y,y')\bigr) \Bigm|  R(x, x') \mbox{ and } S(y, y')\Bigr\} 
\eeq
that is,  treating the two relations independently, and pairing them.

An example when considering {\tt Alice} and {\tt Bob} again is the situation where we know nothing yet about Alice and Bob.  Concretely, where Alice and Bob are both in a space $X$, this is the relation:
\beq
\label{ABunknown0}
\Bigl\{ \bigl( (*,x) , (*,x')  \bigr) \Bigm| x \in X \mbox{ and } x' \in X \Bigr\}
\eeq
This expresses that Alice and Bob could independently be anywhere in $X$. This situation is depicted as follows:
\beq\label{ABunknown}
\input{unknown2.tikz}
\eeq

\paragraph{Composition as a diagram.} Putting sequential and parallel composition together, augmented with caps and cups, we get what are called \emph{string diagrams}.  While these can always be decomposed in terms of the sequential and parallel compositions we have already seen, there is a direct way to compute them without having to do so, for which we refer the reader to Section 3.3.3 of \cite{CKbook}.

An example involving more complex compositions is the equation that justifies representing cap-states and cup-tests as wires:
\[
\input{yank.tikz}
\]
Another one is  the conversion of the different possible representations of  relations mentioned earlier, for example:
\beq\label{eq:CJ}
\input{allR.tikz}
\eeq
So all the ambiguity is now indeed gone, without having to rely on $*$.

\section{A linguistic model of space}  

\subsection{Grammar as relations} 

In the grammars we consider, there are some \emph{basic grammatical types}, like the noun-type $n$, and the type of whole sentences $s$. On the other hand, the transitive-verb-type is not a basic type, but a composite  one made up of two noun-types and a sentence-type.  The precise manner in which these basic types interact depends on which \em grammar calculus \em one uses e.g.~\cite{Ajdukiewicz, Bar-Hillel, Lambek0}.  WAs indicated above, we present our examples in terms of Lambek's \em pregroups \em   \cite{Lambek1, LambekBook}, because of their perfect match with relational diagrams. As shown in \cite{DRichie} there is no loss of generality by choosing pregroup grammars, as they also accommodate CCGs, and do so while maintaining diagrammatic elegance.

For each basic type $x$ there is a left- and a right-inverse ${}^{-1}x$ and~$x^{-1}$.  Then, in English, a transitive verb has type:
\[
{}^{-1}n \cdot s \cdot n^{-1}    
\]
To understand this type, consider a transitive verb like {\tt chases}. Simply saying {\tt chases} doesn't convey any useful information, until we also specify {\tt who  chases who}.  That's the role of the inverses: they specify that in order to form a meaningful sentence  the transitive verb needs a noun on the left and a noun on the right, which then cancel out:   
\[
\underbrace{\mbox{\tt Alice}}_{n} \underbrace{\mbox{\tt chases}}_{{}^{-1}n \cdot s \cdot n^{-1}}  \underbrace{\mbox{\tt Bob}}_n
\]  
So both $n \cdot {}^{-1}n$ and $n^{-1} \cdot n$ vanish, and what remains is $s$, confirming that 
{\tt Alice hates Bob} is a grammatically well-typed sentence. We can depict the cancelations:     
\ctikzfig{hates}
%
%
%
%
%
%
which yields a diagram. For a more complex sentence like: 
\begin{center}
{\tt Alice does not chase Bob} 
\end{center}
the wiring will be more complex \cite{CSC}:
\ctikzfig{hatescomplgram}
but the idea remains the same.  In general, one can extract these wirings from the book \cite{LambekBook}, which assigns types to all grammatical roles. 

The main idea of DisCoCat is to think of these wires not just as a representation of an algebraic computation, but as a representation of how the meanings of the words of the sentence interact.  Since in this paper we are dealing with relations, wires will be made up of cup-relations, e.g.:  
\beq\label{eq:XY}
\Bigl\{\bigl((x,x),*\bigr) \Bigm| x\in X\Bigr\}\times
\Bigl\{\bigl(y,y\bigr) \Bigm| y\in Y\Bigr\}
\times\Bigl\{\bigl((x,x),*\bigr) \Bigm| x\in X\Bigr\}
\eeq
for the 1st example above.  Here $X$ is then the set representing $n$, while $Y$ represents $s$, and we discuss below how $Y$ is expressed in terms of $X$.

\subsection{Combining grammar with spatial meaning}

We now move on to describing our grammatically-structured model of space.  Let space be some set $X$, for example Cartesian space $\mathbb{R}^3$ or space-time, a chessboard $\texttt{a-h} \times \texttt{1-8}$, the metro stations of the $\texttt{Tuen Ma}$ line, or Penrose's staircase $\texttt{I-IV} \times \texttt{1-n}$. The space $X$ is enough to determine the caps/cups, compositions, and shape of spatial relations we have considered above.

A DisCoCat model of `whatever' requires: 
\bit
\item how meanings are represented, also accounting for types, and how we compose meaning spaces in order to allow for multiple words. As we saw above, these word meanings are depicted as states:
\[
\input{hates2.tikz}      
\]
\item a corresponding representation of grammatical structure, as we also saw above:
\[
\input{hatescopy.tikz}   
\]
\item application of grammar to meaning, and we also know how to do that:
\beq\label{hatescopycopy}
\input{hatescopycopy.tikz}    
\eeq   
\eit
One should think of the wires now `feeding' the object and the subject into the verb in order to produce the meaning of the whole sentence. 

\begin{remark}
One concise manner to establish  all the above  is by providing a so-called `compact closed category' (CCC) of meanings \cite{CSC}.  We are fortunate here that for our purposes such a CCC already exists, namely the one of relations with the compositions and caps/cups as described above. We then restrict that category to those objects that we care about.
\end{remark}   
 
\subsection{Internal wirings}\label{sec:internal} 

We now move on to explain how $Y$ relates to $X$ in (\ref{eq:XY}), and hence how sentences $s$ relate to the nouns  $n$ within them. This is not a standard part of grammar calculi, but a feature specifically enabled by DisCoCat/DisCoCirc, used previously in \cite{GrefSadr, KartsaklisSadrzadeh2014, CoeckeText, CoeckeMeich, GramEqs, GramCircs}.

We assume transitive verbs to have the following \emph{internal wiring} \cite{GrefSadr, KartsaklisSadrzadeh2014, CoeckeText}:
\beq\label{s34v2}
\input{s34v2.tikz}   
\eeq   
Using this form, we can compute the spatial meaning of the  sentence {\tt Alice chases Bob} from its words.  If initially we know nothing yet about {\tt Alice} and {\tt Bob}, their states will be as in (\ref{ABunknown}).  So we obtain using (\ref{eq:fusion}, \ref{eq:fusion2}):
\[
\input{with.tikz} 
\]
The general representation of {\tt chases} has been imposed on {\tt Alice} and {\tt Bob}, about whom we now know something: namely that wherever in space(-time) {\tt Alice} is, that spot is in a {\tt chases} relationship with wherever {\tt Bob} is. 

On the other hand, if we already know something about {\tt Alice} and {\tt Bob}, for example their initial separation, then (\ref{hatescopycopy}) will rather look as follows:
\[
\input{hatescopycopycopy.tikz}    
\]   
Filling in (\ref{s34v2}) we then get:
\beq\label{withcopy}
\input{withcopy.tikz}
\eeq
Taking {\tt chases} as in (\ref{ChaseCart}) and the initial relation to be:
\[
\input{AliceBob.tikz} \ := \Bigl\{ x_\texttt{A} = x_\texttt{B} = y_\texttt{A} = y_\texttt{B} = z_\texttt{A} = z_\texttt{B} = 0 \implies t_\texttt{A} = 3 mins, t_\texttt{B} = 0\Bigr\}
\]
 then for (\ref{withcopy}) we get:
 \[
 \Bigl\{\bigl((x,y,z, t),(x,y,z, t') \bigr)\Bigm|t = t' + 3 mins)   \Bigr\}
 \]

Adjectives can be given a similar presentation as (\ref{s34v2}):
\[
\input{adjective.tikz}
\]
and then the analogue to (\ref{withcopy}) would be: 
\beq\label{withcopycopy}
\input{withcopycopy.tikz} 
\eeq
Now we get, not fixing the initial separation:  
\beq\label{withcopycopy2}
\Bigl\{\bigl((x,y,z, t),(x,y,z, t') \bigr)\Bigm|(x,y,z)\in {\tt Paris}\ ,\ \exists \delta t >0 \mbox{ such that } t = t' + \delta t   \Bigr\}
\eeq

Internal structures have also been proposed for relative pronouns \cite{FrobMeanI, FrobMeanII}: 
\[
\input{relpron.tikz}   
\]   
and more recently for a large part of English \cite{GramEqs,GramCircs}, for example:  
\[
\input{nextto.tikz}
\]
Here's an example that involves both adjectives and adpositions. We want to specify the indicated pawn in the following board position using only language referring to other pieces:
\[
\scalebox{0.75}{x
\begin{frame}
\newgame
\chessboard[pgfstyle=straightmove,shortenstart=1ex,arrow=to,linewidth=0.1ex,color=red,shortenstart=1ex,showmover=false, setfen=4r3/2n2k2/P3p1p1/5p2/1P1K3N/2PQ4/r4B2/8 w - - 0 20,
            pgfstyle=border,markfields={g6}] 
\end{frame}
}
\]
There are several pawns on the board, so we have: 
\[ 
\texttt{pawn} = \{\texttt{a6,b4,c3,e6,f5,g6}\} 
\]
We will also make use of {\tt next to} which we saw earlier in (\ref{eq:nextto}).   The pawn happens to be next to a king, but there are two kings:
\[ 
\texttt{king} = \{\texttt{d4, f7}\} 
\]
The noun-phrase {\tt pawn next to a king} has the following diagram: 
\[
\input{PnexttoK.tikz} 
\]
After deforming we get:
\[
\input{PnexttoKbis.tikz}
\]
where {\tt next to a king} is just an application like in (\ref{eq:application}):
\beqa
\input{PnexttoKtrisb.tikz}&=&\input{PnexttoKtrisa.tikz}\\
&=&\Bigl\{(\alpha x, \beta y) \Bigm|   |\alpha - \beta|=1\mbox{ or } |x - y| = 1\Bigr\}\circ \{\texttt{d4, f7}\}\\
&=&\{\texttt{c3,c4,c5,d3,d5,e3,e4,e5,e6,e7,e8,f6,f8,g6,g7,g8}\}
\eeqa
Using (\ref{intersection}) we get the intersection of {\tt the pawn} and {\tt next to a king}:
\[
\scalebox{0.65}{
\begin{frame}
\newgame
\chessboard[pgfstyle=straightmoe,shortenstart=1ex,arrow=to,linewidth=0.1ex,color=red,shortenstart=1ex,showmover=false, setfen=4r3/2n2k2/P3p1p1/5p2/1P1K3N/2PQ4/r4B2/8 w - - 0 20,
		color=green!50,
		colorbackfields={a6,b4,c3,e6,f5,g6}]
\end{frame}
}
\raisebox{1.7cm}{$\ \bigcap\ \ $}  
\scalebox{0.65}{
\begin{frame}
\newgame
\chessboard[pgfstyle=straightmove,shortenstart=1ex,arrow=to,linewidth=0.1ex,color=red,shortenstart=1ex,showmover=false, setfen=4r3/2n2k2/P3p1p1/5p2/1P1K3N/2PQ4/r4B2/8 w - - 0 20,
		color=green!50,
		colorbackfields={c3,c4,c5,d3,d5,e3,e4,e5,e6,e7,e8,f6,f8,g6,g7,g8}]
\end{frame}
}
\raisebox{1.7cm}{$\ =\ \ $} 
\scalebox{0.65}{
\begin{frame}
\newgame
\chessboard[pgfstyle=straightmove,shortenstart=1ex,arrow=to,linewidth=0.1ex,color=red,shortenstart=1ex,showmover=false, setfen=4r3/2n2k2/P3p1p1/5p2/1P1K3N/2PQ4/r4B2/8 w - - 0 20,
		color=green!50,
		colorbackfields={c3,e6,g6}]
\end{frame} 
}
\]
This is not good enough yet, as there are two other pawns that also fit the description, apart from the one we want.  Further specification could be to account the our pawn can be captured by a knight, but this requires accounting for distinct capabilities of different inhabitants, which we address in Section \ref{varyinginhab}, where we also continue this example.

\section{Inferences from spatial language}   

As a first example, we want to see if for diagrams (\ref{withcopycopy}) from relation (\ref{withcopycopy2}) we can infer that also {\tt Bob} is in {\tt Paris}, and this is evidently the case.  We can delete Alice:
\[
\input{inference1.tikz}
\]
and find the following `reduced' state:
\[
{\tt Bob} = \bigl\{(x,y,z, t') \bigm|(x,y,z)\in {\tt Paris}\bigr\} 
\]
Hence, from:
\bit
\item {\tt Alice chases Bob}
\item {\tt Alice is in Paris}
\eit 
we inferred that:
\bit
\item {\tt Bob is in Paris}
\eit
Now consider the sentences:
\bit
\item {\tt The painting is above the chest}
\item {\tt The light is above the painting}
\eit 
which analogous as (\ref{withcopy}) we can represent as a diagrams as follows:
\[
\input{inference2.tikz}
\]
recalling (\ref{above}) where we represented {\tt above} as a relation we obtain:
\[ 
\Bigl\{\bigl((x,y,z),(x',y',z'),(x'',y'',z'') \bigr)\Bigm|  z > z' \ ,\ z'>z'' \Bigr\}
\]
Erasing the painting:
\[
\input{inference3.tikz}
\]
we conclude:
\[ 
\Bigl\{\bigl((x,y,z),(x'',y'',z'') \bigr)\Bigm|  z > z'' \Bigr\}
\]
that is:
\bit
\item {\tt The light is above the chest}
\eit

\begin{remark} 
Given that by Equation \ref{intersection} spiders correspond to \emph{AND}, we can represent `infer' purely graphically. For states $Q$ and $R$, let `from $Q$ we infer $R$' be the equation:
\[
\input{AND1.tikz}\ =\ \input{AND3.tikz}
\]
It is possible to derive the  above conclusion graphically.
\end{remark}

We can also consider a space like the Penrose staircase in linguistic terms, and treat the states as spatial models. For example, consider the sentences:
\bit
\item {\tt II is above I}
\item {\tt III is above II}
\item {\tt IV is above III}
\item {\tt I is above IV}
\eit
Working in a space $X$, if we start from knowing nothing about the positions of \texttt{I}, \texttt{II}, \texttt{III}, \texttt{IV}, after imposing the relations above, we will have a state $X_\texttt{I} \times X_\texttt{II} \times X_\texttt{III} \times X_\texttt{IV}$ encoding all configurations of the four staircases in $X$ consistent with the linguistic description. We represent this as:
\[
\input{penrosecircuit.tikz}
\]
When we take $X$ to be $\mathbb{R}^3$, treating \texttt{above} as in previous examples, we can calculate that the diagram above expresses the empty subset of $\mathbb{R}^3_\texttt{I} \times \mathbb{R}^3_\texttt{II} \times \mathbb{R}^3_\texttt{III} \times \mathbb{R}^3_\texttt{IV}$. This is because there are no height coordinates such that $z^{\tt IV} > z^{\tt III} > z^{\tt II} > z^{\tt I} > z^{\tt IV}$. In other words, given this interpretation of the spatial relation \texttt{above} in $\mathbb{R}^3$ there is no model of the Penrose staircase in physical space: it is an impossible object.

\section{Inhabitants with varying spatial features}\label{varyinginhab}

In an example like chess, when we say {\tt can capture}, this will mean something very different depending on the chesspiece doing the capturing. Generally, different inhabitants have different capabilities and properties in the space they inhabit. This is also the case in Cartesian space, e.g.~how fast different inhabitants can move, e.g.~a cheetah vs.~a slug, and when moving away from representing the inhabitants of Cartesian space as points, then different sizes and shapes will start to play a role as well. 

In order to make this a part of our relations model, we will first augment the representation of the space, besides the locations of the chess board, also accounting for the spatially-related features an inhabitant may have:
\beq\label{AugChess}
\texttt{chessboard} := \left(\texttt{a-h} \times \texttt{1-8}\right) \times \{\sympawn, \symrook,\symbishop, \symknight,\symqueen,\symking\}
\eeq
and e.g.:
\[ 
\texttt{pawn} = \{\texttt{a6,b4,c3,e6,f5,g6}\} \times \{\sympawn\}\qquad\qquad \texttt{king} := \{\texttt{d4,f7}\} \times \{\symking\} 
\]
Similarly, in Cartesian space, we may be consider objects of variable size (a.k.a.~extent) or animals with variable speed, respectively:
\beq\label{animalspace}
\mathbb{R}^3\times\{\raisebox{-1.5mm}{\epsfig{figure=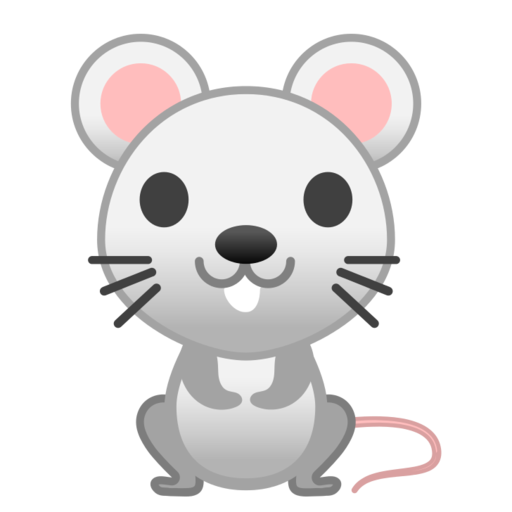,width=14pt}}, \raisebox{-1.5mm}{\epsfig{figure=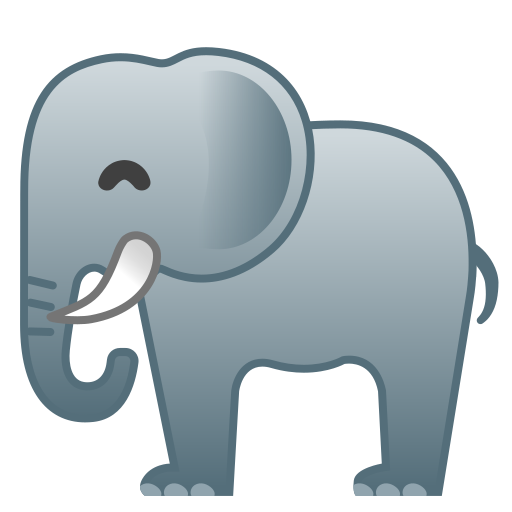,width=14pt}}\} \qquad\qquad\mathbb{R}^3\times\mathbb{R}\times\{ \raisebox{-1.5mm}{\epsfig{figure=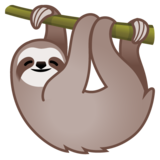,width=14pt}}, \raisebox{-1.5mm}{\epsfig{figure=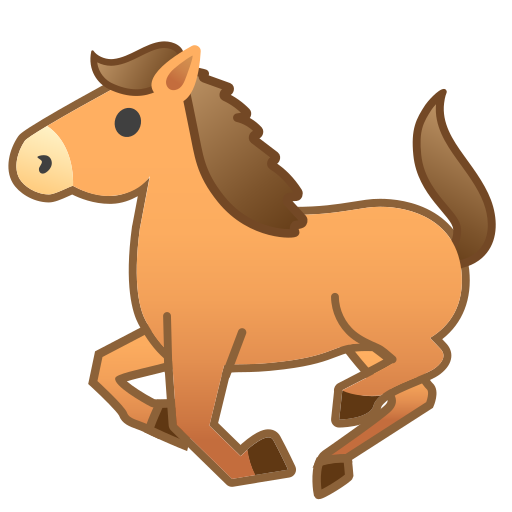,width=14pt}}\}
\eeq
 
\subsection{Chesspieces}

Turning our attention back to the example Section \ref{sec:internal}.  Now consider the following diagram involving {\tt can capture} which requires knowledge about the chesspiece involved: 
\beqa
\input{relpronchessa.tikz}\  &=& \ \input{relpronchessb.tikz}\\
&=& \ \input{relpronchessc.tikz}\\
\eeqa

Now we fine-grain each wire of the {\tt can capture} box as two wires, following equation (\ref{AugChess}): left wire for spatial locations, and right wire for chesspiece labels. The result is a box with four wires as follows:
\[
\input{capturebox.tikz}
\]
We can interpret this relation explicitly as:  
\[
{\tt can\ capture} :=  \left\{
\left(\begin{array}{cc}
\alpha x & {\tt X}\\
\beta y & {\tt Y} 
\end{array}\right)
\Bigm| 
\left(\begin{array}{c} 
\alpha x \\
\beta y 
\end{array}\right)
\in {\tt X's\ moves}\ \mbox{and}\ {\tt X}, {\tt Y}\in \{\sympawn, \symrook,\symbishop, \symknight,\symqueen,\symking\}  \right\} 
\]
where  {\tt X's\ moves} generalises {\tt King's\ moves} in (\ref{kingsmoves}).  Note that {\tt Y} is unconstrained since the kind of the piece being captured plays no role, so the internal structure of the {\tt can capture} box looks as follows: 
\[
{\tt can\ capture} :=\ \input{cancapture1.tikz}   
\]

Above the data of different piece movement rules is encoded within {\tt can\ capture}.  One may argue that it is more natural that, rather than the meaning of {\tt can capture} having to carry the data concerning which pieces can do what, that the chesspieces themselves should carry that data.  This is also possible.  In that case, rather than the space (\ref{AugChess}) we have as the space: 
\[
\texttt{chessboard} := \left(\texttt{a-h} \times \texttt{1-8}\right) \times \{{\tt \sympawn's\ moves}, {\tt \symrook's\ moves}, \ldots, {\tt \symking's\ moves}\}
\]
and:
\[
{\tt can\ capture} :=\ \ \input{cancapture2.tikz}  
\]
since then we have, by (\ref{eq:CJ}): 
\[
\input{cancapture3.tikz}
\]

All together, with our encoding in terms of spatial relations, the noun-phrase: 
\[
\input{relpronchessd.tikz}
\]
now yields the pawn we aimed to characterise, as now we obtain:
\[
\scalebox{0.65}{
\begin{frame}
\newgame
\chessboard[pgfstyle=straightmove,shortenstart=1ex,arrow=to,linewidth=0.1ex,color=red,shortenstart=1ex,showmover=false, setfen=4r3/2n2k2/P3p1p1/5p2/1P1K3N/2PQ4/r4B2/8 w - - 0 20,
		color=green!50,
		colorbackfields={c3,e6,g6}]
\end{frame}
\raisebox{2.7cm}{$\ \bigcap\ \ $}  
}
\scalebox{0.65}{
\begin{frame}
\newgame
\chessboard[pgfstyle=straightmove,shortenstart=1ex,arrow=to,linewidth=0.1ex,color=red,shortenstart=1ex,showmover=false,setfen=4r3/2n2k2/P3p1p1/5p2/1P1K3N/2PQ4/r4B2/8 w - - 0 20,
		pgfstyle=straightmove,
		color=red!50,
		markmoves={c7-a6, h4-f5, h4-g6},
		color=green!50,
		colorbackfields={a6,f5,g6}]
\end{frame}
}
\raisebox{1.7cm}{$\ = \ $}  
\scalebox{0.65}{
\begin{frame}
\newgame
\chessboard[pgfstyle=straightmove,shortenstart=1ex,arrow=to,linewidth=0.1ex,color=red,shortenstart=1ex,showmover=false, setfen=4r3/2n2k2/P3p1p1/5p2/1P1K3N/2PQ4/r4B2/8 w - - 0 20,
		color=green!50,
		colorbackfields={g6}]
\end{frame}
}
\]
Of course, we haven't yet considered that movement is also constrained by whether the destination squares are occupied, and in the case of the Queen, Bishop, and Rook, whether their paths to the destination squares are blocked. Since allowed moves depend on the other pieces on the board, it is arguably more natural to consider collections of pieces, rather than individuals, as inhabitants of the chessboard. Even for the `toy' domain of Chess, there are complicated spatial relations to consider.

\subsection{Capturing in Cartesian space}

Among animals in Cartesian space, keeping things simple, {\tt can capture} is a question of three factors: the distance between hunter and prey, the speeds of each animal, and the endurance of each animal.
Recalling equation (\ref{animalspace}), we might designate this hunting space as:
\[
\texttt{huntspace} := \mathbb{R}^3 \times \underbrace{\mathbb{R} \times \mathbb{R}}_{\!\!\!\!\!\!\!\!\!\text{endurance} \times \text{speed}\!\!\!\!\!\!\!\!\!}
\]
Labelling the wires of the \texttt{can capture} box with the animals' roles, we can depict the internal structure of the \texttt{huntspace} wires as follows:
\[
\input{huntercaptureprey.tikz}\ :=\
\input{capturebox2.tikz}
\]
Explicitly, we might say the \texttt{can capture} relation expresses each animal's running ability as the product of its speed and endurance. The distance between the two animals is considered a head-start for the prey. The hunter can capture the prey when the hunter's running ability exceeds the prey's, accounting for the head-start, and that the duration of the hunt is for as long as the hunter has the endurance to keep hunting.
\[
{\tt can\ capture} :=  \left\{
\left(\begin{array}{ccc}
p_{\tt h} & e_{\tt h} & s_{\tt h}\\
p_{\tt p} & e_{\tt p} & s_{\tt p} 
\end{array}\right)
\Bigm| 
|p_{\tt h} - p_{\tt p}| + \mathop{min}{\{e_{\tt p},e_{\tt h}\}}s_{\tt p} < e_{\tt h}s_{\tt h} \right\} 
\]

Let's say we have the following situation in space, somewhere on the savannah.
\[
\input{savannah0.tikz}
\]
Which ostrich are we referring to, when we say: 
\[
\texttt{The ostrich next to a tree that a cheetah next to grass can capture?}
\]
The language diagram is as follows:
\[
\input{cheetahhuntsentence.tikz}
\]
We can interpret two entities to be \texttt{next to} each other if they are at most, say, 10 metres apart. So we know how to deal with cheetahs next to grass, and ostriches next to trees.
\[
\input{savannahnextto.tikz}
\]
To keep things simple we will ignore acceleration to top speed. Here are the data: 
\bit 
\item \texttt{the cheetah} has a top speed of about 120 kilometres per hour, but an endurance of only 60 seconds. \item \texttt{the ostrich} has a top speed of about 100 kilometres per hour, and they are excellent endurance runners, able to sustain speeds of 50 kilometres per hour for up to 30 minutes. 
\eit
Given our interpretation of \texttt{can capture}, this means cheetahs can catch ostriches unless the ostrich has more than a third of a kilometre head-start. So we know how to concretely evaluate \texttt{can capture} in the space.
\[
\input{savannahcapture.tikz}  
\]
And putting our knowledge together in the same way as in the chesspiece example, we can point out the right ostrich by intersection.
\[
\input{savannah.tikz} 
\]
\par\medskip
\noindent
{\bf Warning.} If you are a cheetah, don't try this at home, as it takes at least three of you to take the ostrich down, and not get killed.    One of us hadn't watched enough nature documentaries before working out this example. As an exercise one can expand this example to account for multiple cheetahs.

\subsection{Extended bodies}

Let's now move away from points to object with a size.  To keep things simple we can assume objects to be spherical, so we only need a radius to specify them.  As an example, let's consider {\tt inside}, for example, fitting a {\tt cheese} inside a {\tt suitcase}.  Then we have:
\[
{\tt inside} :=  \Bigl\{\bigl(((x,y,z), r),((x',y',z'), r') \bigr)\Bigm| r,r' > 0 \mbox{ and } |(x',y',z') - (x,y,z)| < r' - r   \Bigr\} 
\]
which we can now use in a sentence like:
\[
\input{cheese.tikz}
\]
Well, almost, as {\tt stinks} isn't particularly spatial. We now explain how we can combine spatial notions with other concepts that are part of our embodiment.

\section{Concept interaction} 

In \cite{ConcSpacI} a compositional linguistic theory of Gardenfors' conceptual spaces \cite{gardenfors, gardenfors2} was put forward very much along the same lines of our compositional linguistic theory of space.  These conceptual spaces can include spaces of smells, tastes, colours etc.  We can compose our theory with that one, so that we can combine spatial notions with other ones. 

\subsection{Conceptual space meets physical space}  

In conceptual spaces, rather than a plain set, we consider convex sets:
\[
x_1, x_2\in X\ \ \mbox{implies\ $\forall p\in [0,1]$\ that}\ \  p\cdot x_1 + (1-p) \cdot x_2 \in X 
\]
Correspondingly, convex relations satisfy:
 \[
(x_1, y_1), (x_2, y_2)\in R\ \ \mbox{implies\ $\forall p\in [0,1]$\ that}\ \  (p\cdot x_1 + (1-p) \cdot x_2, p\cdot y_1 + (1-p) \cdot y_2)\in R
 \]
For the combined theory, we can consider tuples  of a physical space and one or more conceptual spaces:   
\begin{center}
\epsfig{figure=subway,height=100pt}\ \ \raisebox{1.62cm}{\LARGE$\times$}\ \ \epsfig{figure=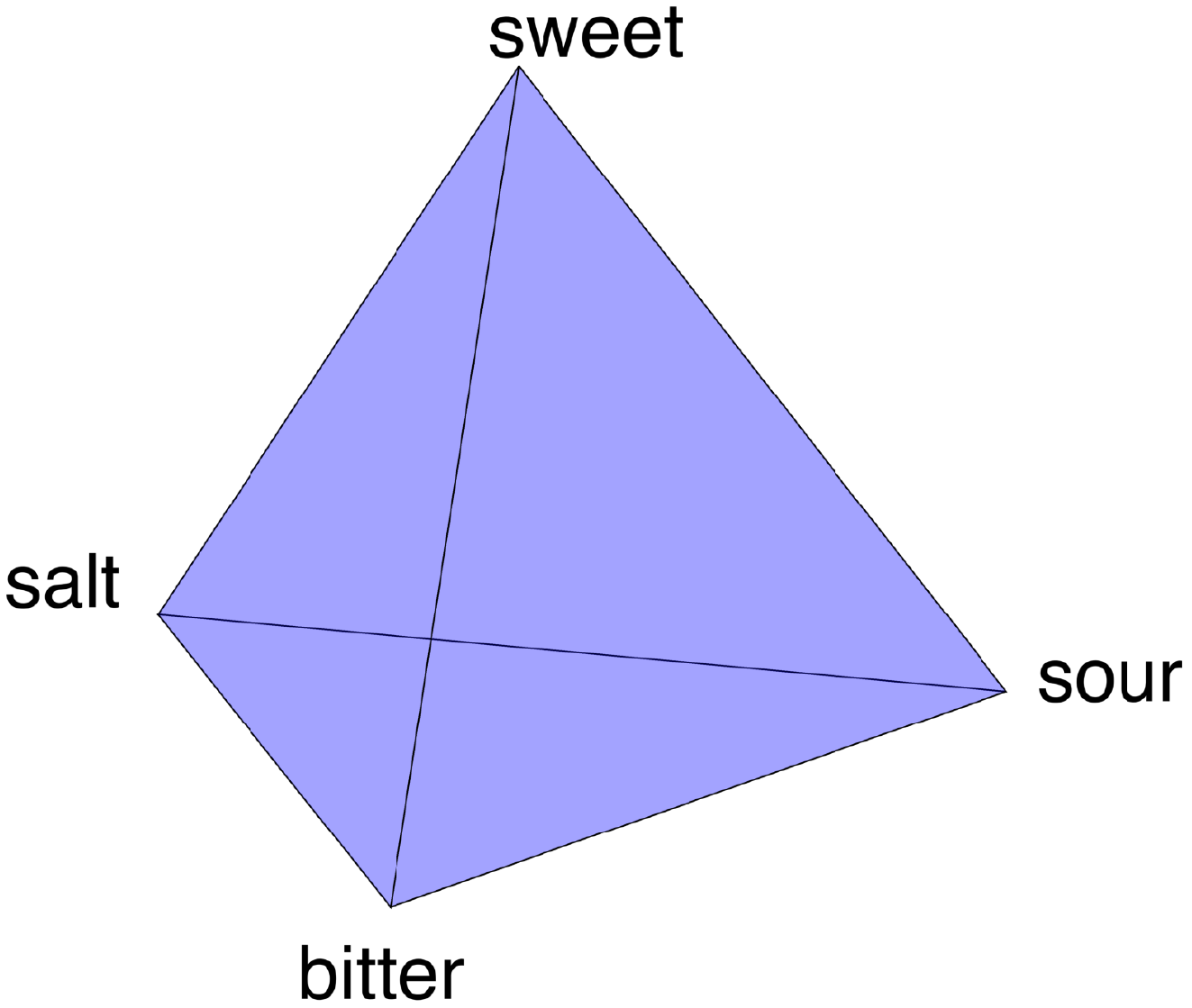,height=100pt}\ \ \raisebox{1.62cm}{\LARGE$\times$}\ \ \epsfig{figure=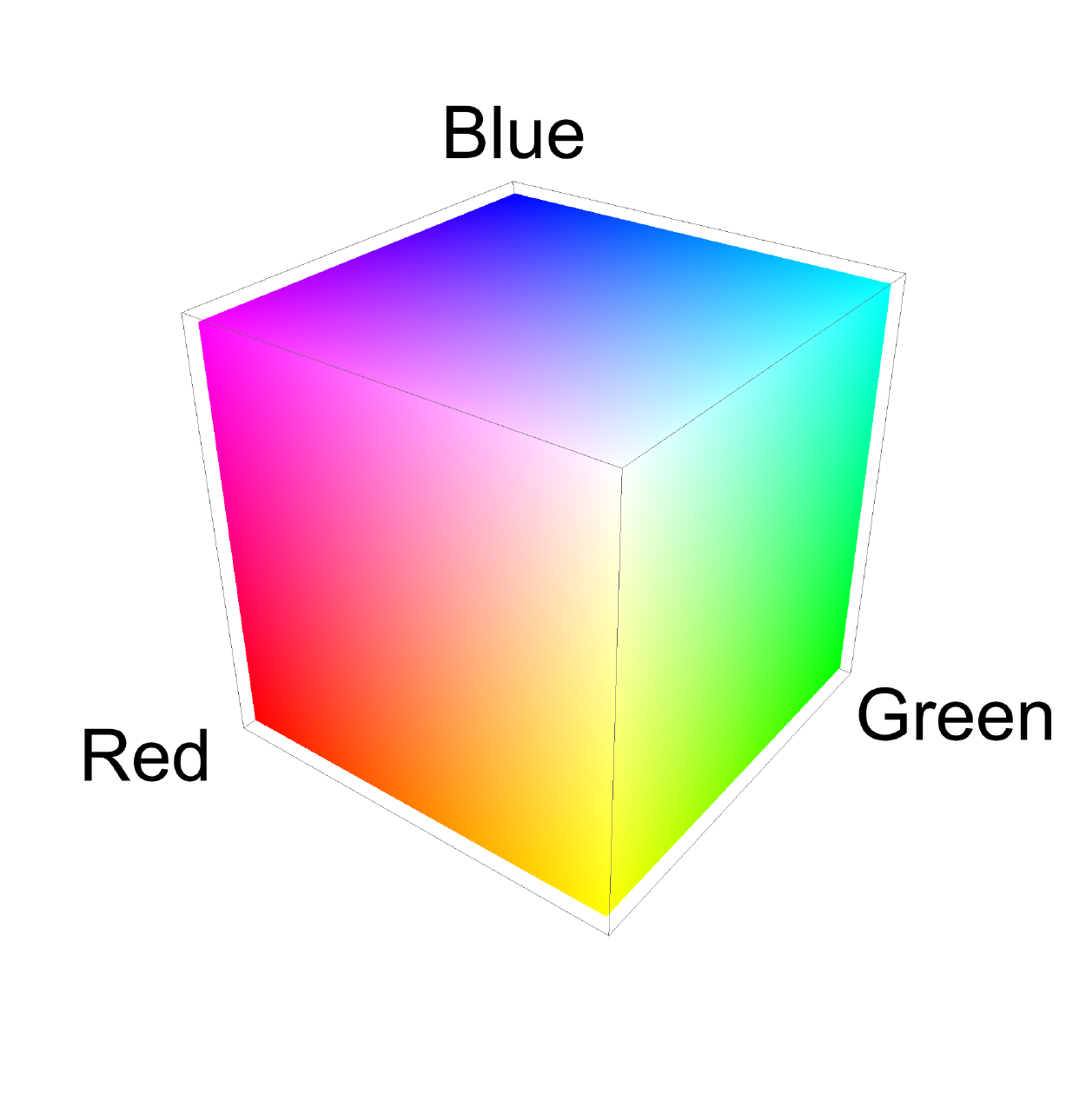,height=100pt}    
\end{center}
The convexity condition for relations on these tuples would then only apply to the convex spaces modelling concepts. On the other hand, while it does not capture all spatial notions, we could also adopt  convexity-compliant models of spatial prepositions -- as G\"{a}rdenfors does in \cite{gardenfors_geometry_2015} -- in order to obtain a compressed theory that could feasibly support empirical experiments.

\subsection{The cheese inside the suitcase stinks}

We need two spaces to deal with this sentence. The first is some formal space of fragrances. The other kind of space we deal with in this example is the physical space of size and extent, $\mathbb{R}^3 \times \mathbb{R}$. So the compound space we are considering is:
\[
\mathbb{R}^3 \times \mathbb{R} \times {\tt fragrance}
\]
We assert that, in addition to physical location and extent, \texttt{the cheese} has a fragrance, which we depict with a dashed wire:
\[
\input{cheesestate.tikz}
\]
As a relation in physical space, \texttt{inside} relates physical positions and extents, but otherwise leaves smell data alone. We depict this as `passing on' the fragrance wire without modifying it, which clearly generalises to other `ignored' spaces.  The reverse order of the wires on the left are to distinguish inputs from outputs when we eventually revisit the language diagram:
\[
\input{cheeseinsidestate.tikz} 
\]
\texttt{the suitcase} has a physical position and extent, but unlike cheese, we do not expect suitcases to smell:
\[
\input{cheesesuitcasestate.tikz}
\]
\texttt{stinks} as an adjective affects the fragrance wire alone, analogous to how \texttt{inside} only wants to deal with the wires of physical space:
\[
\input{cheesestinksstate.tikz}
\]

Now we can revisit the old language diagram. Replacing the boxes we had before:  
\[
\input{cheese2.tikz}
\]
Straightening the wires, we can see clearly two facts about \texttt{cheese}, one about it's relations in physical space, and one regarding its fragrance space:
\[
\input{cheese3.tikz}
\]
So we have -- independently of how we model fragrance-space and physical space -- from \texttt{the cheese inside the suitcase stinks}, that:

\bit
\item {\tt The cheese is inside the suitcase}
\item {\tt The cheese stinks}
\eit 

%

\section{Discussion and Prospects}

\subsection{Relationship to Montague semantics}\label{app}  

Much of what we say here is not specific to this paper, but also applies to other DisCo-related papers, such as \cite{DBLP:conf/wollic/CoeckeGLM17, coecke2018uniqueness, ConcSpacI, tull2020monoidal}.  Since the very first DisCoCat papers \cite{CCS, CSC} it  has been informally claimed that  when DisCoCat-meanings are taken in the category of states and relations this generates a fragment of Montague semantics, something that has been further substantiated in \cite{DBLP:journals/corr/abs-1905-07408}.  So what is the precise relationship between DisCoCat models and Montague semantics?

Firstly, the term Montague semantics has been used for a wide range of different things.  We outline three (successively more general) conceptions, that broadly cover what might be considered Montague semantics. There is Montague's own work \cite{montague1970universal, montague1973proper}, which is the mapping of  categorial grammars, either by direct homomorphism or indirectly via lambda calculus, to  formulas in a particular system of higher-order intensional logic {\bf (M1)}. It has been used to refer to -- e.g.~in \cite{janssen1986foundations} -- \emph{any} structure-preserving map from a grammar into any semantics representable by sets and relations {\bf (M2)}, or even any map of grammar into meaning, where the grammatical structure determines the structure of meaning representations {\bf (M3)}.

Concerning {\bf (M1)} and {\bf (M2)}, Montague semantics have clear truth-theoretic underpinnings, while most DisCoCat models do not. Ultimately, the point of Montague semantics  on these accounts  is that, interpreted in a certain world, a logical statement holds.  On the other hand, DisCoCat meanings aim to `paint a picture' of a certain statement, e.g.~a cheetah chasing an ostrich should be understood as the picture that \emph{shows} this event happening, rather than an \emph{evaluation} that this is the case. Part of these pictures are the specific properties agents in the space have, as discussed in section \ref{varyinginhab}. This difference in aim is what caused distributional semantics to start living a separate existence from semantics \`{a} la Montague \cite{Gazdar, ClarkPulman}, and it was DisCoCat that brought them back together \cite{CCS, CSC}. 

Concerning {\bf (M3)}, modern NLP has clearly demonstrated that grammar cannot live an independent existence from meaning, so taking  meaning  as the target of a mapping is problematic.  More recent DisCoCat approaches are gradually becoming a hybrid that combines grammar and meaning on equal footing, so they can mutually inform each other.  One feature of this blurring of the distinction between grammar and meaning are the internal wirings of Section \ref{sec:internal}.

\subsection{Prospects}

Imagine a computer learning and computing with concepts: how far off are we?

Concepts, whether spatio-temporal or more general, are productive and systematic \cite{fodor_concepts_1998} in a similar way that language is. There are no dictionaries for sentences, but sentences are -- to a certain extent, excluding contextual content -- systematically produced by composites of individual words that can be formally captured in a dictionary. Similarly, there are no complete gallery collections of concepts, but complex concepts can be painted by formal palettes of simpler concepts \cite{gardenfors}. Without making any claims regarding the nature of concepts, nor about how human cognitive architectures work, we suggest it is worthwhile to design a program that leverages the systematic and productive capacity of language to structure concept acquisition and formation. We suggest it is worthwhile to bring the machinery of DisCoCirc to bear on this difficult and well-studied problem.

Naturally, there are directions in which the theory must extend to fit practical purposes. The concepts we have presented here, as relations, are binary things: every element is either completely in or out. A probabilistic and fuzzy picture of linguistically-compliant concepts \cite{vincent_wang_concept_2020,tull_categorical_2021} will be more useful in practice, and even necessary if the practical aim is to compute with human-friendly concepts. The differentiable structure of fuzzy concepts is also crucially necessary in order to allow concepts to be learned from data, rather than axiomatised and handcrafted `bottom-up'. Efforts in that direction are currently underway.  

There are endless further aspects of the structure of concepts, such as hyponymy \cite{bankova2019graded,MarthaDot}, embedding in spacetime and causal structure \cite{signorelli_cognitive_2020}, the role of metaphor in shaping concepts \cite{lakoff_metaphors_2003,hofstadter_surfaces_2013}, to name a few. There is no claim these formal theories capture empirically how \emph{humans} think, but we can still evaluate these frameworks by whether they can be incorporated into practical computational tools.

\section{Acknowledgements}  

We thank Sean Tull,  Steve Clark, Alexis Toumi and Ilyas Khan for very useful discussions about the content of this paper. We thank Michael Moortgat for useful ideas on a previous version of the paper.  This challenge to have a linguistic theory about agents in space stretches back many years to initial discussions with Dan Marsden and Robin Piedeleu.

\bibliographystyle{plain}
\bibliography{mainNOW}

\end{document}

%% file: macros/defs.tex
\newkeycommand{\boxmap}[style=small box][1]{\,\tikz{\node[style=\commandkey{style}] (x) {$#1$};\draw(0,-1.25)--(x)--(0,1.25);}\,}
\newkeycommand{\dboxmap}[style=dbox][1]{\,\tikz{\node[style=\commandkey{style}] (x) {$#1$};\draw[boldedge] (0,-1.25)--(x)--(0,1.25);}\,}

\newkeycommand{\wirelabel}[style=white label]{\,\tikz{\node[style=\commandkey{style}]{};}\,\xspace}

\newkeycommand{\pointmap}[style=point][1]{\,\tikz{\node[style=\commandkey{style}] (x) at (0,-0.3) {$#1$};\node [style=none] (3) at (0, 0.9) {};\node [style=none] (3) at (0, -0.9) {};\draw (x)--(0,0.7);}\,}

\newkeycommand{\point}[style=point,estyle=][1]{\,\tikz{\node[style=\commandkey{style}] (x) at (0,-0.05) {$#1$};\draw [\commandkey{estyle}] (x)--(0,1.05);}\,}

\newkeycommand{\copointmap}[style=copoint][1]{\,\tikz{\node[style=\commandkey{style}] (x) at (0,0.3) {$#1$};\draw (x)--(0,-0.7);}\,}

\newkeycommand{\dpoint}[style=point][1]{\,\tikz{\node[style=\commandkey{style},doubled] (x) at (0,-0.3) {$#1$};\draw[boldedge] (x)--(0,0.7);}\,}

\newkeycommand{\kpoint}[style=kpoint,estyle=][1]{\,\tikz{\node[style=\commandkey{style}] (x) at (0,-0.05) {$#1$};\draw [\commandkey{estyle}] (x)--(0,1.05);}\,}

\newkeycommand{\kpointgray}[style=gray kpoint,estyle=][1]{\,\tikz{\node[style=\commandkey{style}] (x) at (0,-0.05) {$#1$};\draw [\commandkey{estyle}] (x)--(0,1.05);}\,}

\newkeycommand{\typedkpoint}[style=kpoint,edgestyle=][2]{%
\,\begin{tikzpicture}
  \begin{pgfonlayer}{nodelayer}
    \node [style=none] (0) at (0, 1) {};
    \node [style=\commandkey{style}] (1) at (0, -0.75) {$#2$};
    \node [style=right label] (2) at (0.25, 0.5) {$#1$};
  \end{pgfonlayer}
  \begin{pgfonlayer}{edgelayer}
    \draw [\commandkey{edgestyle}] (1) to (0.center);
  \end{pgfonlayer}
\end{tikzpicture}\,}

\newkeycommand{\typedkpointdag}[style=kpoint adjoint,edgestyle=][2]{%
\,\begin{tikzpicture}
  \begin{pgfonlayer}{nodelayer}
    \node [style=none] (0) at (0, -1) {};
    \node [style=\commandkey{style}] (1) at (0, 0.75) {$#2$};
    \node [style=right label] (2) at (0.25, -0.5) {$#1$};
  \end{pgfonlayer}
  \begin{pgfonlayer}{edgelayer}
    \draw [\commandkey{edgestyle}] (0.center) to (1);
  \end{pgfonlayer}
\end{tikzpicture}\,}

\newkeycommand{\bistate}[style=kpoint][1]{\,\begin{tikzpicture}
  \begin{pgfonlayer}{nodelayer}
    \node [style=\commandkey{style}, minimum width=1 cm, inner sep=2pt] (0) at (0, -0.25) {$#1$};
    \node [style=none] (1) at (-0.75, 0) {};
    \node [style=none] (2) at (0.75, 0) {};
    \node [style=none] (3) at (-0.75, 0.88) {};
    \node [style=none] (4) at (0.75, 0.88) {};
  \end{pgfonlayer}
  \begin{pgfonlayer}{edgelayer}
    \draw (1.center) to (3.center);
    \draw (2.center) to (4.center);
  \end{pgfonlayer}
\end{tikzpicture}\,}

\newkeycommand{\bistatebig}[style=kpoint][1]{\,\begin{tikzpicture}
  \begin{pgfonlayer}{nodelayer}
    \node [style=\commandkey{style}, minimum width=, minimum width=1.26 cm, inner sep=2pt] (0) at (0, -0.25) {$#1$};
    \node [style=none] (1) at (-1, 0) {};
    \node [style=none] (2) at (1, 0) {};
    \node [style=none] (3) at (-1, 0.88) {};
    \node [style=none] (4) at (1, 0.88) {};
  \end{pgfonlayer}
  \begin{pgfonlayer}{edgelayer}
    \draw (1.center) to (3.center);
    \draw (2.center) to (4.center);
  \end{pgfonlayer}
\end{tikzpicture}\,}

\newkeycommand{\dbistate}[style=dkpoint][1]{\,\begin{tikzpicture}
  \begin{pgfonlayer}{nodelayer}
    \node [style=\commandkey{style}, minimum width=1 cm, inner sep=2pt] (0) at (0, -0.25) {$#1$};
    \node [style=none] (1) at (-0.75, 0) {};
    \node [style=none] (2) at (0.75, 0) {};
    \node [style=none] (3) at (-0.75, 1.0) {};
    \node [style=none] (4) at (0.75, 1.0) {};
  \end{pgfonlayer}
  \begin{pgfonlayer}{edgelayer}
    \draw [boldedge] (1.center) to (3.center);
    \draw [boldedge] (2.center) to (4.center);
  \end{pgfonlayer}
\end{tikzpicture}\,}

\newkeycommand{\bistatebraket}[style1=kpoint,style2=kpointadj][2]{\,\begin{tikzpicture}
  \begin{pgfonlayer}{nodelayer}
    \node [style=\commandkey{style1}, minimum width=1 cm, inner sep=2pt] (0) at (0, -0.75) {$#1$};
    \node [style=none] (1) at (-0.75, -0.5) {};
    \node [style=none] (2) at (0.75, -0.5) {};
    \node [style=none] (3) at (-0.75, 0.5) {};
    \node [style=none] (4) at (0.75, 0.5) {};
    \node [style=\commandkey{style2}, minimum width=1 cm, inner sep=2pt] (5) at (0, 0.75) {$#2$};
  \end{pgfonlayer}
  \begin{pgfonlayer}{edgelayer}
    \draw (1.center) to (3.center);
    \draw (2.center) to (4.center);
  \end{pgfonlayer}
\end{tikzpicture}\,}

\newkeycommand{\weight}[style=dkpoint][1]{\,\begin{tikzpicture}
  \begin{pgfonlayer}{nodelayer}
    \node [style=\commandkey{style}] (0) at (0, -0.5) {$#1$};
    \node [style=upground] (1) at (0, 0.75) {};
  \end{pgfonlayer}
  \begin{pgfonlayer}{edgelayer}
    \draw [style=boldedge] (0) to (1);
  \end{pgfonlayer}
\end{tikzpicture}\,}

\newkeycommand{\dkpoint}[style=dkpoint][1]{\,\tikz{\node[style=\commandkey{style}] (x) at (0,-0.1) {$#1$};\draw[boldedge] (x)--(0,1);}\,}
\newkeycommand{\dkpointadj}[style=dkpointadj][1]{\,\tikz{\node[style=\commandkey{style}] (x) at (0,0.1) {$#1$};\draw[boldedge] (x)--(0,-1);}\,}
\newkeycommand{\dkpointtrans}[style=dkpointtrans][1]{\,\tikz{\node[style=\commandkey{style}] (x) at (0,0.1) {$#1$};\draw[boldedge] (x)--(0,-1);}\,}
\newkeycommand{\dkpointconj}[style=dkpointconj][1]{\,\tikz{\node[style=\commandkey{style}] (x) at (0,-0.1) {$#1$};\draw[boldedge] (x)--(0,1);}\,}

\newkeycommand{\blackkpoint}[style=black kpoint][1]{\,\tikz{\node[style=\commandkey{style}] (x) at (0,-0.1) {$#1$};\draw (x)--(0,1);}\,}
\newkeycommand{\blackkpointadj}[style=black kpointadj][1]{\,\tikz{\node[style=\commandkey{style}] (x) at (0,0.1) {$#1$};\draw (x)--(0,-1);}\,}
\newkeycommand{\blackdkpoint}[style=black dkpoint][1]{\,\tikz{\node[style=\commandkey{style}] (x) at (0,-0.1) {$#1$};\draw[boldedge] (x)--(0,1);}\,}
\newkeycommand{\blackdkpointadj}[style=black dkpointadj][1]{\,\tikz{\node[style=\commandkey{style}] (x) at (0,0.1) {$#1$};\draw[boldedge] (x)--(0,-1);}\,}

\newcommand{\trace}{\,\begin{tikzpicture}
  \begin{pgfonlayer}{nodelayer}
    \node [style=none] (0) at (0, -0.60) {};
    \node [style=upground] (1) at (0, 0.40) {};
  \end{pgfonlayer}
  \begin{pgfonlayer}{edgelayer}
    \draw [style=none] (0.center) to (1);
  \end{pgfonlayer}
\end{tikzpicture}\,}

\newcommand\discard\trace

\newkeycommand{\pointketbra}[style1=copoint,style2=point,estyle=][2]{\,%
\begin{tikzpicture}
  \begin{pgfonlayer}{nodelayer}
    \node [style=none] (0) at (0, -1.75) {};
    \node [style=\commandkey{style1}] (1) at (0, -0.75) {$#1$};
    \node [style=\commandkey{style2}] (2) at (0, 0.75) {$#2$};
    \node [style=none] (3) at (0, 1.75) {};
  \end{pgfonlayer}
  \begin{pgfonlayer}{edgelayer}
    \draw [\commandkey{estyle}] (0.center) to (1);
    \draw [\commandkey{estyle}] (2) to (3.center);
  \end{pgfonlayer}
\end{tikzpicture}\,}

\newkeycommand{\twocopointketbra}[style1=copoint,style2=copoint,style3=point][3]{\,%
\begin{tikzpicture}
  \begin{pgfonlayer}{nodelayer}
    \node [style=none] (0) at (-0.75, -1.75) {};
    \node [style=none] (1) at (0.75, -1.75) {};
    \node [style=\commandkey{style1}] (2) at (-0.75, -0.75) {$#1$};
    \node [style=\commandkey{style2}] (3) at (0.75, -0.75) {$#2$};
    \node [style=\commandkey{style3}] (4) at (0, 0.75) {$#3$};
    \node [style=none] (5) at (0, 1.75) {};
  \end{pgfonlayer}
  \begin{pgfonlayer}{edgelayer}
    \draw (0.center) to (2);
    \draw (1.center) to (3);
    \draw (4) to (5.center);
  \end{pgfonlayer}
\end{tikzpicture}\,}

\newkeycommand{\twopointketbra}[style1=copoint,style2=point,style3=point][3]{\,%
\begin{tikzpicture}
  \begin{pgfonlayer}{nodelayer}
    \node [style=none] (0) at (0, -1.75) {};
    \node [style=none] (1) at (-0.75, 1.75) {};
    \node [style=\commandkey{style1}] (2) at (0, -0.75) {$#1$};
    \node [style=\commandkey{style2}] (3) at (-0.75, 0.75) {$#2$};
    \node [style=\commandkey{style3}] (4) at (0.75, 0.75) {$#3$};
    \node [style=none] (5) at (0.75, 1.75) {};
  \end{pgfonlayer}
  \begin{pgfonlayer}{edgelayer}
    \draw (0.center) to (2);
    \draw (1.center) to (3);
    \draw (4) to (5.center);
  \end{pgfonlayer}
\end{tikzpicture}\,}

\newkeycommand{\pointbraket}[style1=point,style2=copoint,estyle=][2]{\,%
\begin{tikzpicture}
  \begin{pgfonlayer}{nodelayer}
    \node [style=\commandkey{style1}] (0) at (0, -0.625) {$#1$};
    \node [style=\commandkey{style2}] (1) at (0, 0.625) {$#2$};
  \end{pgfonlayer}
  \begin{pgfonlayer}{edgelayer}
    \draw [\commandkey{estyle}] (0) to (1);
  \end{pgfonlayer}
\end{tikzpicture}\,}

\newkeycommand{\kpointketbra}[style1=kpointdag,style2=kpoint,estyle=][2]{\,%
\begin{tikzpicture}
  \begin{pgfonlayer}{nodelayer}
    \node [style=none] (0) at (0, -1.9) {};
    \node [style=\commandkey{style1}] (1) at (0, -0.95) {$#1$};
    \node [style=\commandkey{style2}] (2) at (0, 0.95) {$#2$};
    \node [style=none] (3) at (0, 1.9) {};
  \end{pgfonlayer}
  \begin{pgfonlayer}{edgelayer}
    \draw [\commandkey{estyle}] (0.center) to (1);
    \draw [\commandkey{estyle}] (2) to (3.center);
  \end{pgfonlayer}
\end{tikzpicture}\,}

\newkeycommand{\kpointmap}[style1=kpoint,style2=map][2]{\,%
\begin{tikzpicture}
  \begin{pgfonlayer}{nodelayer}
    \node [style=\commandkey{style1}] (0) at (0, -1.1) {$#1$};
    \node [style=\commandkey{style2}] (1) at (0, 0.5) {$#2$};
    \node [style=none] (2) at (0, 1.75) {};
  \end{pgfonlayer}
  \begin{pgfonlayer}{edgelayer}
    \draw (0) to (1);
    \draw (1) to (2.center);
  \end{pgfonlayer}
\end{tikzpicture}%
\,}

\newkeycommand{\kpointmaptrans}[style1=kpointconj,style2=mapadj][2]{\,%
\begin{tikzpicture}
  \begin{pgfonlayer}{nodelayer}
    \node [style=\commandkey{style1}] (0) at (0, -1.1) {$#1$};
    \node [style=\commandkey{style2}] (1) at (0, 0.5) {$#2$};
    \node [style=none] (2) at (0, 1.75) {};
  \end{pgfonlayer}
  \begin{pgfonlayer}{edgelayer}
    \draw (0) to (1);
    \draw (1) to (2.center);
  \end{pgfonlayer}
\end{tikzpicture}%
\,}

\newkeycommand{\kpointmapadj}[style1=kpointadj,style2=map][2]{\,%
\begin{tikzpicture}
  \begin{pgfonlayer}{nodelayer}
    \node [style=\commandkey{style1}] (0) at (0, 1.1) {$#1$};
    \node [style=\commandkey{style2}] (1) at (0, -0.5) {$#2$};
    \node [style=none] (2) at (0, -1.75) {};
  \end{pgfonlayer}
  \begin{pgfonlayer}{edgelayer}
    \draw (0) to (1);
    \draw (1) to (2.center);
  \end{pgfonlayer}
\end{tikzpicture}%
\,}

\newkeycommand{\dkpointmap}[style1=dkpoint,style2=dmap][2]{\,%
\begin{tikzpicture}
  \begin{pgfonlayer}{nodelayer}
    \node [style=\commandkey{style1}] (0) at (0, -1.2) {$#1$};
    \node [style=\commandkey{style2}] (1) at (0, 0.4) {$#2$};
    \node [style=none] (2) at (0, 1.7) {};
  \end{pgfonlayer}
  \begin{pgfonlayer}{edgelayer}
    \draw[style=boldedge] (0) to (1);
    \draw[style=boldedge] (1) to (2.center);
  \end{pgfonlayer}
\end{tikzpicture}%
\,}

\newkeycommand{\dkpointmapx}[style1=dkpoint,style2=dmap][2]{\,%
\begin{tikzpicture}
  \begin{pgfonlayer}{nodelayer}
    \node [style=\commandkey{style1}] (0) at (0, -1.4) {$#1$};
    \node [style=\commandkey{style2}] (1) at (0, 0.6) {$#2$};
    \node [style=none] (2) at (0, 1.9) {};
  \end{pgfonlayer}
  \begin{pgfonlayer}{edgelayer}
    \draw[style=boldedge] (0) to (1);
    \draw[style=boldedge] (1) to (2.center);
  \end{pgfonlayer}
\end{tikzpicture}%
\,}

\newkeycommand{\boxcopointmapdag}[style1=map,style2=copoint][2]{\,%
\begin{tikzpicture}
  \begin{pgfonlayer}{nodelayer}
    \node [style=none] (0) at (0, -1.75) {};
    \node [style=\commandkey{style1}] (1) at (0, -0.5) {$#1$};
    \node [style=\commandkey{style2}] (2) at (0, 1) {$#2$};
  \end{pgfonlayer}
  \begin{pgfonlayer}{edgelayer}
    \draw (0.center) to (1);
    \draw (1) to (2);
  \end{pgfonlayer}
\end{tikzpicture}%
\,}

\newkeycommand{\kpointdagmap}[style1=map,style2=kpointdag][2]{\,%
\begin{tikzpicture}
  \begin{pgfonlayer}{nodelayer}
    \node [style=none] (0) at (0, -1.75) {};
    \node [style=\commandkey{style1}] (1) at (0, -0.5) {$#1$};
    \node [style=\commandkey{style2}] (2) at (0, 1.1) {$#2$};
  \end{pgfonlayer}
  \begin{pgfonlayer}{edgelayer}
    \draw (0.center) to (1);
    \draw (1) to (2);
  \end{pgfonlayer}
\end{tikzpicture}%
\,}

\newkeycommand{\sandwichmap}[style1=point,style2=map,style3=copoint][3]{\,%
\begin{tikzpicture}
  \begin{pgfonlayer}{nodelayer}
    \node [style=\commandkey{style1}] (0) at (0, -1.50) {$#1$};
    \node [style=\commandkey{style2}] (1) at (0, 0) {$#2$};
    \node [style=\commandkey{style3}] (2) at (0, 1.50) {$#3$};
  \end{pgfonlayer}
  \begin{pgfonlayer}{edgelayer}
    \draw (0) to (1);
    \draw (1) to (2);
  \end{pgfonlayer}
\end{tikzpicture}%
}

\newkeycommand{\kpointsandwichmap}[style1=kpoint,style2=map,style3=kpointdag][3]{\,%
\begin{tikzpicture}
  \begin{pgfonlayer}{nodelayer}
    \node [style=\commandkey{style1}] (0) at (0, -1.5) {$#1$};
    \node [style=\commandkey{style2}] (1) at (0, 0) {$#2$};
    \node [style=\commandkey{style3}] (2) at (0, 1.5) {$#3$};
  \end{pgfonlayer}
  \begin{pgfonlayer}{edgelayer}
    \draw (0) to (1);
    \draw (1) to (2);
  \end{pgfonlayer}
\end{tikzpicture}%
}

\newkeycommand{\sandwichmapconj}[style1=point,style2=mapconj,style3=copoint][3]{\,%
\begin{tikzpicture}
  \begin{pgfonlayer}{nodelayer}
    \node [style=\commandkey{style1}] (0) at (0, -1.50) {$#1$};
    \node [style=\commandkey{style2}] (1) at (0, 0) {$#2$};
    \node [style=\commandkey{style3}] (2) at (0, 1.50) {$#3$};
  \end{pgfonlayer}
  \begin{pgfonlayer}{edgelayer}
    \draw (0) to (1);
    \draw (1) to (2);
  \end{pgfonlayer}
\end{tikzpicture}%
}

\newkeycommand{\sandwichtwo}[style1=point,style2=map,style3=map,style4=copoint][4]{\,%
\begin{tikzpicture}
  \begin{pgfonlayer}{nodelayer}
    \node [style=\commandkey{style2}] (0) at (0, -1) {$#2$};
    \node [style=\commandkey{style3}] (1) at (0, 1) {$#3$};
    \node [style=\commandkey{style1}] (2) at (0, -2.6) {$#1$};
    \node [style=\commandkey{style4}] (3) at (0, 2.6) {$#4$};
  \end{pgfonlayer}
  \begin{pgfonlayer}{edgelayer}
    \draw (2) to (0);
    \draw (0) to (1);
    \draw (1) to (3);
  \end{pgfonlayer}
\end{tikzpicture}\,}

\newkeycommand{\longbraket}[style1=point,style2=copoint][2]{\,%
\begin{tikzpicture}
  \begin{pgfonlayer}{nodelayer}
    \node [style=\commandkey{style1}] (0) at (0, -2.6) {$#1$};
    \node [style=\commandkey{style2}] (1) at (0, 2.6) {$#2$};
  \end{pgfonlayer}
  \begin{pgfonlayer}{edgelayer}
    \draw (0) to (1);
  \end{pgfonlayer}
\end{tikzpicture}\,}

\newkeycommand{\onb}[style=point]{\ensuremath{\{\,\tikz{\node[style=\commandkey{style}] (x) at (0,-0.3) {$j$};\draw (x)--(0,0.7);}\,\}}\xspace}

\newkeycommand{\phase}[style=white phase dot][1]{\,\begin{tikzpicture}
    \begin{pgfonlayer}{nodelayer}
        \node [style=none] (0) at (0, 1) {};
        \node [style=\commandkey{style}] (2) at (0, -0) {$#1$}; 
        \node [style=none] (3) at (0, -1) {};
    \end{pgfonlayer}
    \begin{pgfonlayer}{edgelayer}
        \draw (2) to (0.center);
        \draw (3.center) to (2);
    \end{pgfonlayer}
\end{tikzpicture}\,}

\newkeycommand{\dphase}[style=white phase ddot][1]{\,\begin{tikzpicture}
    \begin{pgfonlayer}{nodelayer}
        \node [style=none] (0) at (0, 1.25) {};
        \node [style=\commandkey{style}] (2) at (0, -0) {$#1$};
        \node [style=none] (3) at (0, -1.25) {};
    \end{pgfonlayer}
    \begin{pgfonlayer}{edgelayer}
        \draw [boldedge] (2) to (0.center);
        \draw [boldedge] (3.center) to (2);
    \end{pgfonlayer}
\end{tikzpicture}\,}

\newkeycommand{\dphasepoint}[style=white phase ddot][1]{\,\begin{tikzpicture}[yshift=-3mm]
  \begin{pgfonlayer}{nodelayer}
    \node [style=none] (0) at (0, 1) {};
    \node [style=\commandkey{style}] (1) at (0, 0) {$#1$};
  \end{pgfonlayer}
  \begin{pgfonlayer}{edgelayer}
    \draw [style=boldedge] (0.center) to (1);
  \end{pgfonlayer}
\end{tikzpicture}\,}

\newkeycommand{\dphasepointgray}[style=gray phase ddot][1]{\,\begin{tikzpicture}[yshift=-3mm]
  \begin{pgfonlayer}{nodelayer}
    \node [style=none] (0) at (0, 1) {};
    \node [style=\commandkey{style}] (1) at (0, 0) {$#1$};
  \end{pgfonlayer}
  \begin{pgfonlayer}{edgelayer}
    \draw [style=boldedge] (0.center) to (1);
  \end{pgfonlayer}
\end{tikzpicture}\,}

\newkeycommand{\dphasecopoint}[style=white phase ddot][1]{\,\begin{tikzpicture}[yshift=3mm,yscale=-1]
  \begin{pgfonlayer}{nodelayer}
    \node [style=none] (0) at (0, 1) {};
    \node [style=\commandkey{style}] (1) at (0, 0) {$#1$};
  \end{pgfonlayer}
  \begin{pgfonlayer}{edgelayer}
    \draw [style=boldedge] (0.center) to (1);
  \end{pgfonlayer}
\end{tikzpicture}\,}

\newkeycommand{\dphasepointsm}[style=white phase ddot][1]{\,\begin{tikzpicture}[yshift=-3mm]
  \begin{pgfonlayer}{nodelayer}
    \node [style=none] (0) at (0, 1) {};
    \node [style=\commandkey{style}] (1) at (0, 0) {\footnotesize\!$#1$\!};
  \end{pgfonlayer}
  \begin{pgfonlayer}{edgelayer}
    \draw [style=boldedge] (0.center) to (1);
  \end{pgfonlayer}
\end{tikzpicture}\,}

\newkeycommand{\phasepoint}[style=white phase dot][1]{\,\begin{tikzpicture}[yshift=-3mm]
  \begin{pgfonlayer}{nodelayer}
    \node [style=none] (0) at (0, 1) {};
    \node [style=\commandkey{style}] (1) at (0, 0) {$#1$};
  \end{pgfonlayer}
  \begin{pgfonlayer}{edgelayer}
    \draw (0.center) to (1);
  \end{pgfonlayer}
\end{tikzpicture}\,}

\newkeycommand{\phasecopoint}[style=white phase dot][1]{\,\begin{tikzpicture}[yshift=3mm]
  \begin{pgfonlayer}{nodelayer}
    \node [style=none] (0) at (0, -1) {};
    \node [style=\commandkey{style}] (1) at (0, 0) {$#1$};
  \end{pgfonlayer}
  \begin{pgfonlayer}{edgelayer}
    \draw (0.center) to (1);
  \end{pgfonlayer}
\end{tikzpicture}\,}

\newkeycommand{\kpointbraket}[style1=kpoint,style2=kpoint adjoint,estyle=][2]{\,%
\begin{tikzpicture}
  \begin{pgfonlayer}{nodelayer}
    \node [style=\commandkey{style1}] (0) at (0, -0.735) {$#1$};
    \node [style=\commandkey{style2}] (1) at (0, 0.735) {$#2$};
  \end{pgfonlayer}
  \begin{pgfonlayer}{edgelayer}
    \draw [\commandkey{estyle}] (0) to (1);
  \end{pgfonlayer}
\end{tikzpicture}\,}

\newkeycommand{\kkpointbraket}[style1=kpoint,style2=kpoint adjoint,estyle=][2]{\,%
\begin{tikzpicture}
  \begin{pgfonlayer}{nodelayer}
    \node [style=\commandkey{style1}] (0) at (0, -0.685) {$#1$};
    \node [style=\commandkey{style2}] (1) at (0, 0.685) {$#2$};
  \end{pgfonlayer}
  \begin{pgfonlayer}{edgelayer}
    \draw [\commandkey{estyle}] (0) to (1);
  \end{pgfonlayer}
\end{tikzpicture}\,}

\newkeycommand{\kpointbraketx}[style1=kpoint,style2=kpoint adjoint,estyle=][2]{\,%
\begin{tikzpicture}
  \begin{pgfonlayer}{nodelayer}
    \node [style=\commandkey{style1}] (0) at (0, -0.885) {$#1$};
    \node [style=\commandkey{style2}] (1) at (0, 0.885) {$#2$};
  \end{pgfonlayer}
  \begin{pgfonlayer}{edgelayer}
    \draw [\commandkey{estyle}] (0) to (1);
  \end{pgfonlayer}
\end{tikzpicture}\,}

\tikzstyle{cdiag}=[matrix of math nodes, row sep=3em, column sep=3em, text height=1.5ex, text depth=0.25ex,inner sep=0.5em]
\tikzstyle{arrow above}=[transform canvas={yshift=0.5ex}]
\tikzstyle{arrow below}=[transform canvas={yshift=-0.5ex}]



%% file: macros/tikzstyles.tex
\usepackage{tikz}
\usetikzlibrary{decorations.markings}
\usetikzlibrary{shapes.geometric}

\pgfdeclarelayer{edgelayer}
\pgfdeclarelayer{nodelayer}
\pgfsetlayers{edgelayer,nodelayer,main}

\tikzstyle{none}=[inner sep=0pt]
\definecolor{hexcolor0xff0000}{rgb}{1.000,0.000,0.000}
\definecolor{hexcolor0x000000}{rgb}{0.000,0.000,0.000}
\definecolor{hexcolor0x00ff00}{rgb}{0.000,1.000,0.000}
\definecolor{hexcolor0x000000}{rgb}{0.000,0.000,0.000}
\definecolor{hexcolor0xffff00}{rgb}{1.000,1.000,0.000}
\definecolor{hexcolor0xffffff}{rgb}{1.000,1.000,1.000}

\tikzstyle{rn}=[circle,fill=hexcolor0xff0000,draw=hexcolor0x000000,line width=0.8 pt]
\tikzstyle{gn}=[circle,fill=hexcolor0x00ff00,draw=hexcolor0x000000,line width=0.8 pt]
\tikzstyle{yn}=[circle,fill=hexcolor0xffff00,draw=hexcolor0x000000,line width=0.8 pt]
\tikzstyle{wn}=[circle,fill=hexcolor0xffffff,draw=hexcolor0x000000,line width=0.8 pt]
\tikzstyle{wnthick}=[circle,fill=hexcolor0xffffff,draw=hexcolor0x000000,line width=2.500]

\tikzstyle{simple}=[-,draw=hexcolor0x000000,line width=2.000]
\tikzstyle{arrow}=[-,draw=hexcolor0x000000,postaction={decorate},decoration={markings,mark=at position .5 with {\arrow{>}}},line width=2.000]
\tikzstyle{tick}=[-,draw=hexcolor0x000000,postaction={decorate},decoration={markings,mark=at position .5 with {\draw (0,-0.1) -- (0,0.1);}},line width=2.000]
\tikzstyle{halfthickness}=[-,draw=hexcolor0x000000,line width=0.500]
\tikzstyle{thick}=[-,draw=hexcolor0x000000,line width=2.500]
\tikzstyle{thicker}=[-,draw=hexcolor0x000000,line width=4.000]

\tikzstyle{env}=[copoint,regular polygon rotate=0,minimum width=0.2cm, fill=black]

\tikzstyle{probs}=[shape=semicircle,fill=white,draw=black,shape border rotate=180,minimum width=1.2cm]

\tikzstyle{every picture}=[baseline=-0.25em,scale=0.5]
\tikzstyle{dotpic}=[] 
\tikzstyle{diredges}=[every to/.style={diredge}]
\tikzstyle{math matrix}=[matrix of math nodes,left delimiter=(,right delimiter=),inner sep=2pt,column sep=1em,row sep=0.5em,nodes={inner sep=0pt},text height=1.5ex, text depth=0.25ex]

\tikzstyle{inline text}=[text height=1.5ex, text depth=0.25ex,yshift=0.5mm]
\tikzstyle{label}=[font=\footnotesize,text height=1.5ex, text depth=0.25ex,yshift=0.5mm]
\tikzstyle{left label}=[label,anchor=east,xshift=1.5mm]
\tikzstyle{right label}=[label,anchor=west,xshift=-1.5mm]

\tikzstyle{braceedge}=[decorate,decoration={brace,amplitude=2mm,raise=-1mm}]
\tikzstyle{small braceedge}=[decorate,decoration={brace,amplitude=1mm,raise=-1mm}]

\tikzstyle{doubled}=[line width=1.6pt] 
\tikzstyle{boldedge}=[doubled,shorten <=-0.17mm,shorten >=-0.17mm]
\tikzstyle{boldedgegray}=[doubled,gray,shorten <=-0.17mm,shorten >=-0.17mm]

\tikzstyle{semidoubled}=[line width=1.4pt] 
\tikzstyle{semiboldedgegray}=[semidoubled,gray,shorten <=-0.17mm,shorten >=-0.17mm]

\tikzstyle{boldedgedashed}=[very thick,dashed,shorten <=-0.17mm,shorten >=-0.17mm]
\tikzstyle{vboldedgedashed}=[doubled,dashed,shorten <=-0.17mm,shorten >=-0.17mm]
\tikzstyle{left hook arrow}=[left hook-latex]
\tikzstyle{right hook arrow}=[right hook-latex]
\tikzstyle{sembracket}=[line width=0.5pt,shorten <=-0.07mm,shorten >=-0.07mm]

\tikzstyle{causal edge}=[->,thick,gray]
\tikzstyle{causal nondir}=[thick,gray]
\tikzstyle{timeline}=[thick,gray, dashed]

\tikzstyle{cedge}=[<->,thick,gray!70!white]

\tikzstyle{empty diagram}=[draw=gray!40!white,dashed,shape=rectangle,minimum width=1cm,minimum height=1cm]
\tikzstyle{empty diagram small}=[draw=gray!50!white,dashed,shape=rectangle,minimum width=0.6cm,minimum height=0.5cm]

\tikzstyle{dot}=[inner sep=0mm,minimum width=2mm,minimum height=2mm,draw,shape=circle]
\tikzstyle{ddot}=[inner sep=0mm, doubled, minimum width=2.5mm,minimum height=2.5mm,draw,shape=circle]

\tikzstyle{black dot}=[dot,fill=black]
\tikzstyle{white dot}=[dot,fill=white,,text depth=-0.2mm]
\tikzstyle{green dot}=[white dot] 
\tikzstyle{gray dot}=[dot,fill=gray!40!white,,text depth=-0.2mm]
\tikzstyle{red dot}=[gray dot] 

\tikzstyle{black ddot}=[ddot,fill=black]
\tikzstyle{white ddot}=[ddot,fill=white]
\tikzstyle{gray ddot}=[ddot,fill=gray!40!white]

\tikzstyle{gray edge}=[gray!40!white]

\tikzstyle{small dot}=[inner sep=0.3mm,minimum width=0pt,minimum height=0pt,draw,shape=circle]

\tikzstyle{small black dot}=[small dot,fill=black]
\tikzstyle{small white dot}=[small dot,fill=white]
\tikzstyle{small gray dot}=[small dot,fill=gray!40!white]

\tikzstyle{causal dot}=[inner sep=0.4mm,minimum width=0pt,minimum height=0pt,draw=white,shape=circle,fill=gray!40!white]

\tikzstyle{phase dimensions}=[minimum size=5mm,font=\footnotesize,rectangle,rounded corners=2.5mm,inner sep=0.2mm,outer sep=-2mm]
\tikzstyle{dphase dimensions}=[minimum size=5mm,font=\footnotesize,rectangle,rounded corners=2.5mm,inner sep=0.2mm,outer sep=-2mm]

\tikzstyle{white phase dot}=[dot,fill=white,phase dimensions]
\tikzstyle{white phase ddot}=[ddot,fill=white,dphase dimensions]
\tikzstyle{green phase ddot}=[ddot,fill=green,dphase dimensions]

\tikzstyle{white rect ddot}=[draw=black,fill=white,doubled,minimum size=5mm,font=\footnotesize,rectangle,rounded corners=2.5mm,inner sep=0.2mm]
\tikzstyle{gray rect ddot}=[draw=black,fill=gray!40!white,doubled,minimum size=6mm,font=\footnotesize,rectangle,rounded corners=3mm]

\tikzstyle{gray phase dot}=[dot,fill=gray!40!white,phase dimensions]
\tikzstyle{gray phase ddot}=[ddot,fill=gray!40!white,dphase dimensions]
\tikzstyle{red phase ddot}=[ddot,fill=red,dphase dimensions]

\tikzstyle{grey phase dot}=[gray phase dot]
\tikzstyle{grey phase ddot}=[gray phase ddot]

\tikzstyle{small phase dimensions}=[minimum size=4mm,font=\tiny,rectangle,rounded corners=2mm,inner sep=0.2mm,outer sep=-2mm]
\tikzstyle{small dphase dimensions}=[minimum size=4mm,font=\tiny,rectangle,rounded corners=2mm,inner sep=0.2mm,outer sep=-2mm]

\tikzstyle{small gray phase dot}=[dot,fill=gray!40!white,small phase dimensions]
\tikzstyle{small gray phase ddot}=[ddot,fill=gray!40!white,small dphase dimensions]

\tikzstyle{small map}=[draw,shape=rectangle,minimum height=4mm,minimum width=4mm,fill=white]

\tikzstyle{cnot}=[fill=white,shape=circle,inner sep=-1.4pt]

\tikzstyle{asym hadamard}=[fill=white,draw,shape=NEbox,inner sep=0.6mm,font=\footnotesize,minimum height=4mm]
\tikzstyle{asym hadamard conj}=[fill=white,draw,shape=NWbox,inner sep=0.6mm,font=\footnotesize,minimum height=4mm]
\tikzstyle{asym hadamard dag}=[fill=white,draw,shape=SEbox,inner sep=0.6mm,font=\footnotesize,minimum height=4mm]

\tikzstyle{hadamard}=[fill=white,draw,inner sep=0.6mm,font=\footnotesize,minimum height=4mm,minimum width=4mm]
\tikzstyle{small hadamard}=[fill=white,draw,inner sep=0.6mm,minimum height=1.5mm,minimum width=1.5mm]
\tikzstyle{dhadamard}=[hadamard,doubled]
\tikzstyle{small dhadamard}=[small hadamard,doubled]
\tikzstyle{small dhadamard rotate}=[small hadamard,doubled,rotate=45]
\tikzstyle{antipode}=[white dot,inner sep=0.3mm,font=\footnotesize]

\tikzstyle{scalar}=[diamond,draw,inner sep=0.5pt,font=\small]
\tikzstyle{dscalar}=[diamond,doubled, draw,inner sep=0.5pt,font=\small]

\tikzstyle{small box}=[rectangle,inline text,fill=white,draw,minimum height=5mm,yshift=-0.5mm,minimum width=5mm,font=\small]
\tikzstyle{small gray box}=[small box,fill=gray!30]
\tikzstyle{medium box}=[rectangle,inline text,fill=white,draw,minimum height=5mm,yshift=-0.5mm,minimum width=10mm,font=\small]
\tikzstyle{square box}=[small box] 
\tikzstyle{medium gray box}=[small box,fill=gray!30]
\tikzstyle{semilarge box}=[rectangle,inline text,fill=white,draw,minimum height=5mm,yshift=-0.5mm,minimum width=12.5mm,font=\small]
\tikzstyle{large box}=[rectangle,inline text,fill=white,draw,minimum height=5mm,yshift=-0.5mm,minimum width=15mm,font=\small]
\tikzstyle{large gray box}=[small box,fill=gray!30]

\tikzstyle{Bayes box}=[rectangle,fill=black,draw, minimum height=3mm, minimum width=3mm]

\tikzstyle{gray square point}=[small box,fill=gray!50]

\tikzstyle{dphase box white}=[dhadamard]
\tikzstyle{dphase box gray}=[dhadamard,fill=gray!50!white]

\tikzstyle{point}=[regular polygon,regular polygon sides=3,draw,scale=0.75,inner sep=-0.5pt,minimum width=9mm,fill=white,regular polygon rotate=180]
\tikzstyle{copoint}=[regular polygon,regular polygon sides=3,draw,scale=0.75,inner sep=-0.5pt,minimum width=9mm,fill=white]
\tikzstyle{dpoint}=[point,doubled]
\tikzstyle{dcopoint}=[copoint,doubled]

\tikzstyle{wide copoint}=[fill=white,draw,shape=isosceles triangle,shape border rotate=90,isosceles triangle stretches=true,inner sep=0pt,minimum width=1.5cm,minimum height=6.12mm]
\tikzstyle{wide point}=[fill=white,draw,shape=isosceles triangle,shape border rotate=-90,isosceles triangle stretches=true,inner sep=0pt,minimum width=1.5cm,minimum height=6.12mm,yshift=-0.0mm]
\tikzstyle{wide point plus}=[fill=white,draw,shape=isosceles triangle,shape border rotate=-90,isosceles triangle stretches=true,inner sep=0pt,minimum width=1.74cm,minimum height=7mm,yshift=-0.0mm]

\tikzstyle{wide dpoint}=[fill=white,doubled,draw,shape=isosceles triangle,shape border rotate=-90,isosceles triangle stretches=true,inner sep=0pt,minimum width=1.5cm,minimum height=6.12mm,yshift=-0.0mm]
\tikzstyle{wide dcopoint}=[fill=white,doubled,draw,shape=isosceles triangle,shape border rotate=90,isosceles triangle stretches=true,inner sep=0pt,minimum width=1.5cm,minimum height=6.12mm,yshift=-0.0mm]

\tikzstyle{tinypoint}=[regular polygon,regular polygon sides=3,draw,scale=0.55,inner sep=-0.15pt,minimum width=6mm,fill=white,regular polygon rotate=180]

\tikzstyle{white point}=[point]
\tikzstyle{white dpoint}=[dpoint]
\tikzstyle{green point}=[white point] 
\tikzstyle{white copoint}=[copoint]
\tikzstyle{gray point}=[point,fill=gray!40!white]
\tikzstyle{gray dpoint}=[gray point,doubled]
\tikzstyle{red point}=[gray point] 
\tikzstyle{gray copoint}=[copoint,fill=gray!40!white]
\tikzstyle{gray dcopoint}=[gray copoint,doubled]

\tikzstyle{white point guide}=[regular polygon,regular polygon sides=3,font=\scriptsize,draw,scale=0.65,inner sep=-0.5pt,minimum width=9mm,fill=white,regular polygon rotate=180]

\tikzstyle{black point}=[point,fill=black,font=\color{white}]
\tikzstyle{black copoint}=[copoint,fill=black,font=\color{white}]

\tikzstyle{tiny gray point}=[tinypoint,fill=gray!40!white]

\tikzstyle{diredge}=[->]
\tikzstyle{ddiredge}=[<->]
\tikzstyle{rdiredge}=[<-]
\tikzstyle{thickdiredge}=[->, very thick]
\tikzstyle{pointer edge}=[->,very thick,gray]
\tikzstyle{pointer edge part}=[very thick,gray]
\tikzstyle{dashed edge}=[dashed]
\tikzstyle{thick dashed edge}=[very thick,dashed]
\tikzstyle{thick gray dashed edge}=[thick dashed edge,gray!40]
\tikzstyle{thick map edge}=[very thick,|->]

\makeatletter
\newcommand{\boxshape}[3]{%
\pgfdeclareshape{#1}{
\inheritsavedanchors[from=rectangle] 
\inheritanchorborder[from=rectangle]
\inheritanchor[from=rectangle]{center}
\inheritanchor[from=rectangle]{north}
\inheritanchor[from=rectangle]{south}
\inheritanchor[from=rectangle]{west}
\inheritanchor[from=rectangle]{east}
\backgroundpath{
\southwest \pgf@xa=\pgf@x \pgf@ya=\pgf@y
\northeast \pgf@xb=\pgf@x \pgf@yb=\pgf@y

\@tempdima=#2
\@tempdimb=#3

\pgfpathmoveto{\pgfpoint{\pgf@xa - 5pt + \@tempdima}{\pgf@ya}}
\pgfpathlineto{\pgfpoint{\pgf@xa - 5pt - \@tempdima}{\pgf@yb}}
\pgfpathlineto{\pgfpoint{\pgf@xb + 5pt + \@tempdimb}{\pgf@yb}}
\pgfpathlineto{\pgfpoint{\pgf@xb + 5pt - \@tempdimb}{\pgf@ya}}
\pgfpathlineto{\pgfpoint{\pgf@xa - 5pt + \@tempdima}{\pgf@ya}}
\pgfpathclose
}
}}

\boxshape{NEbox}{0pt}{5pt}
\boxshape{SEbox}{0pt}{-5pt}
\boxshape{NWbox}{5pt}{0pt}
\boxshape{SWbox}{-5pt}{0pt}
\boxshape{EBox}{-3pt}{3pt}
\boxshape{WBox}{3pt}{-3pt}
\makeatother

\tikzstyle{cloud}=[shape=cloud,draw,minimum width=1.5cm,minimum height=1.5cm]

\tikzstyle{map}=[draw,shape=NEbox,inner sep=2pt,minimum height=6mm,fill=white]
\tikzstyle{dashedmap}=[draw,dashed,shape=NEbox,inner sep=2pt,minimum height=6mm,fill=white]
\tikzstyle{mapdag}=[draw,shape=SEbox,inner sep=2pt,minimum height=6mm,fill=white]
\tikzstyle{mapadj}=[draw,shape=SEbox,inner sep=2pt,minimum height=6mm,fill=white]
\tikzstyle{maptrans}=[draw,shape=SWbox,inner sep=2pt,minimum height=6mm,fill=white]
\tikzstyle{mapconj}=[draw,shape=NWbox,inner sep=2pt,minimum height=6mm,fill=white]

\tikzstyle{langmap}=[draw,shape=NEbox,inner sep=2pt,minimum height=2.4mm,minimum width=3.2mm,fill=white]
\tikzstyle{langmaptrans}=[draw,shape=SWbox,inner sep=2pt,minimum height=2.4mm,minimum width=3.2mm,fill=white]

\tikzstyle{medium map}=[draw,shape=NEbox,inner sep=2pt,minimum height=6mm,fill=white,minimum width=7mm]
\tikzstyle{medium map dag}=[draw,shape=SEbox,inner sep=2pt,minimum height=6mm,fill=white,minimum width=7mm]
\tikzstyle{medium map adj}=[draw,shape=SEbox,inner sep=2pt,minimum height=6mm,fill=white,minimum width=7mm]
\tikzstyle{medium map trans}=[draw,shape=SWbox,inner sep=2pt,minimum height=6mm,fill=white,minimum width=7mm]
\tikzstyle{medium map conj}=[draw,shape=NWbox,inner sep=2pt,minimum height=6mm,fill=white,minimum width=7mm]
\tikzstyle{semilarge map}=[draw,shape=NEbox,inner sep=2pt,minimum height=6mm,fill=white,minimum width=9.5mm]
\tikzstyle{semilarge map trans}=[draw,shape=SWbox,inner sep=2pt,minimum height=6mm,fill=white,minimum width=9.5mm]
\tikzstyle{semilarge map adj}=[draw,shape=SEbox,inner sep=2pt,minimum height=6mm,fill=white,minimum width=9.5mm]
\tikzstyle{semilarge map dag}=[draw,shape=SEbox,inner sep=2pt,minimum height=6mm,fill=white,minimum width=9.5mm]
\tikzstyle{semilarge map conj}=[draw,shape=NWbox,inner sep=2pt,minimum height=6mm,fill=white,minimum width=9.5mm]
\tikzstyle{large map}=[draw,shape=NEbox,inner sep=2pt,minimum height=6mm,fill=white,minimum width=12mm]
\tikzstyle{large map conj}=[draw,shape=NWbox,inner sep=2pt,minimum height=6mm,fill=white,minimum width=12mm]
\tikzstyle{very large map}=[draw,shape=NEbox,inner sep=2pt,minimum height=6mm,fill=white,minimum width=17mm]

\tikzstyle{medium dmap}=[draw,doubled,shape=NEbox,inner sep=2pt,minimum height=6mm,fill=white,minimum width=7mm]
\tikzstyle{medium dmap dag}=[draw,doubled,shape=SEbox,inner sep=2pt,minimum height=6mm,fill=white,minimum width=7mm]
\tikzstyle{medium dmap adj}=[draw,doubled,shape=SEbox,inner sep=2pt,minimum height=6mm,fill=white,minimum width=7mm]
\tikzstyle{medium dmap trans}=[draw,doubled,shape=SWbox,inner sep=2pt,minimum height=6mm,fill=white,minimum width=7mm]
\tikzstyle{medium dmap conj}=[draw,doubled,shape=NWbox,inner sep=2pt,minimum height=6mm,fill=white,minimum width=7mm]
\tikzstyle{semilarge dmap}=[draw,doubled,shape=NEbox,inner sep=2pt,minimum height=6mm,fill=white,minimum width=9.5mm]
\tikzstyle{semilarge dmap trans}=[draw,doubled,shape=SWbox,inner sep=2pt,minimum height=6mm,fill=white,minimum width=9.5mm]
\tikzstyle{semilarge dmap adj}=[draw,doubled,shape=SEbox,inner sep=2pt,minimum height=6mm,fill=white,minimum width=9.5mm]
\tikzstyle{semilarge dmap dag}=[draw,doubled,shape=SEbox,inner sep=2pt,minimum height=6mm,fill=white,minimum width=9.5mm]
\tikzstyle{semilarge dmap conj}=[draw,doubled,shape=NWbox,inner sep=2pt,minimum height=6mm,fill=white,minimum width=9.5mm]
\tikzstyle{large dmap}=[draw,doubled,shape=NEbox,inner sep=2pt,minimum height=6mm,fill=white,minimum width=12mm]
\tikzstyle{large dmap conj}=[draw,doubled,shape=NWbox,inner sep=2pt,minimum height=6mm,fill=white,minimum width=12mm]
\tikzstyle{large dmap trans}=[draw,doubled,shape=SWbox,inner sep=2pt,minimum height=6mm,fill=white,minimum width=12mm]
\tikzstyle{large dmap adj}=[draw,doubled,shape=SEbox,inner sep=2pt,minimum height=6mm,fill=white,minimum width=12mm]
\tikzstyle{large dmap dag}=[draw,doubled,shape=SEbox,inner sep=2pt,minimum height=6mm,fill=white,minimum width=12mm]
\tikzstyle{very large dmap}=[draw,doubled,shape=NEbox,inner sep=2pt,minimum height=6mm,fill=white,minimum width=19.5mm]

\tikzstyle{muxbox}=[draw,shape=rectangle,minimum height=3mm,minimum width=3mm,fill=white]
\tikzstyle{dmuxbox}=[muxbox,doubled]

\tikzstyle{box}=[draw,shape=rectangle,inner sep=2pt,minimum height=6mm,minimum width=6mm,fill=white]
\tikzstyle{dbox}=[draw,doubled,shape=rectangle,inner sep=2pt,minimum height=6mm,minimum width=6mm,fill=white]
\tikzstyle{dmap}=[draw,doubled,shape=NEbox,inner sep=2pt,minimum height=6mm,fill=white]
\tikzstyle{dmapdag}=[draw,doubled,shape=SEbox,inner sep=2pt,minimum height=6mm,fill=white]
\tikzstyle{dmapadj}=[draw,doubled,shape=SEbox,inner sep=2pt,minimum height=6mm,fill=white]
\tikzstyle{dmaptrans}=[draw,doubled,shape=SWbox,inner sep=2pt,minimum height=6mm,fill=white]
\tikzstyle{dmapconj}=[draw,doubled,shape=NWbox,inner sep=2pt,minimum height=6mm,fill=white]

\tikzstyle{ddmap}=[draw,doubled,dashed,shape=NEbox,inner sep=2pt,minimum height=6mm,fill=white]
\tikzstyle{ddmapdag}=[draw,doubled,dashed,shape=SEbox,inner sep=2pt,minimum height=6mm,fill=white]
\tikzstyle{ddmapadj}=[draw,doubled,dashed,shape=SEbox,inner sep=2pt,minimum height=6mm,fill=white]
\tikzstyle{ddmaptrans}=[draw,doubled,dashed,shape=SWbox,inner sep=2pt,minimum height=6mm,fill=white]
\tikzstyle{ddmapconj}=[draw,doubled,dashed,shape=NWbox,inner sep=2pt,minimum height=6mm,fill=white]

\boxshape{sNEbox}{0pt}{3pt}
\boxshape{sSEbox}{0pt}{-3pt}
\boxshape{sNWbox}{3pt}{0pt}
\boxshape{sSWbox}{-3pt}{0pt}
\tikzstyle{smap}=[draw,shape=sNEbox,fill=white]
\tikzstyle{smapdag}=[draw,shape=sSEbox,fill=white]
\tikzstyle{smapadj}=[draw,shape=sSEbox,fill=white]
\tikzstyle{smaptrans}=[draw,shape=sSWbox,fill=white]
\tikzstyle{smapconj}=[draw,shape=sNWbox,fill=white]

\tikzstyle{dsmap}=[draw,dashed,shape=sNEbox,fill=white]
\tikzstyle{dsmapdag}=[draw,dashed,shape=sSEbox,fill=white]
\tikzstyle{dsmaptrans}=[draw,dashed,shape=sSWbox,fill=white]
\tikzstyle{dsmapconj}=[draw,dashed,shape=sNWbox,fill=white]

\boxshape{mNEbox}{0pt}{10pt}
\boxshape{mSEbox}{0pt}{-10pt}
\boxshape{mNWbox}{10pt}{0pt}
\boxshape{mSWbox}{-10pt}{0pt}
\tikzstyle{mmap}=[draw,shape=mNEbox]
\tikzstyle{mmapdag}=[draw,shape=mSEbox]
\tikzstyle{mmaptrans}=[draw,shape=mSWbox]
\tikzstyle{mmapconj}=[draw,shape=mNWbox]

\tikzstyle{mmapgray}=[draw,fill=gray!40!white,shape=mNEbox]
\tikzstyle{smapgray}=[draw,fill=gray!40!white,shape=sNEbox]

\makeatletter
\pgfdeclareshape{cornerpoint}{
\inheritsavedanchors[from=rectangle] 
\inheritanchorborder[from=rectangle]
\inheritanchor[from=rectangle]{center}
\inheritanchor[from=rectangle]{north}
\inheritanchor[from=rectangle]{south}
\inheritanchor[from=rectangle]{west}
\inheritanchor[from=rectangle]{east}
\backgroundpath{
\southwest \pgf@xa=\pgf@x \pgf@ya=\pgf@y
\northeast \pgf@xb=\pgf@x \pgf@yb=\pgf@y

\pgfmathsetmacro{\pgf@shorten@left}{\pgfkeysvalueof{/tikz/shorten left}}
\pgfmathsetmacro{\pgf@shorten@right}{\pgfkeysvalueof{/tikz/shorten right}}

\pgfpathmoveto{\pgfpoint{0.5 * (\pgf@xa + \pgf@xb)}{\pgf@ya - 5pt}}
\pgfpathlineto{\pgfpoint{\pgf@xa - 8pt + \pgf@shorten@left}{\pgf@yb - 1.5 * \pgf@shorten@left}}
\pgfpathlineto{\pgfpoint{\pgf@xa - 8pt + \pgf@shorten@left}{\pgf@yb}}
\pgfpathlineto{\pgfpoint{\pgf@xb + 8pt - \pgf@shorten@right}{\pgf@yb}}
\pgfpathlineto{\pgfpoint{\pgf@xb + 8pt - \pgf@shorten@right}{\pgf@yb - 1.5 * \pgf@shorten@right}}
\pgfpathclose
}
}

\pgfdeclareshape{cornercopoint}{
\inheritsavedanchors[from=rectangle] 
\inheritanchorborder[from=rectangle]
\inheritanchor[from=rectangle]{center}
\inheritanchor[from=rectangle]{north}
\inheritanchor[from=rectangle]{south}
\inheritanchor[from=rectangle]{west}
\inheritanchor[from=rectangle]{east}
\backgroundpath{
\southwest \pgf@xa=\pgf@x \pgf@ya=\pgf@y
\northeast \pgf@xb=\pgf@x \pgf@yb=\pgf@y

\pgfmathsetmacro{\pgf@shorten@left}{\pgfkeysvalueof{/tikz/shorten left}}
\pgfmathsetmacro{\pgf@shorten@right}{\pgfkeysvalueof{/tikz/shorten right}}

\pgfpathmoveto{\pgfpoint{0.5 * (\pgf@xa + \pgf@xb)}{\pgf@yb + 5pt}}
\pgfpathlineto{\pgfpoint{\pgf@xa - 8pt + \pgf@shorten@left}{\pgf@ya + 1.5 * \pgf@shorten@left}}
\pgfpathlineto{\pgfpoint{\pgf@xa - 8pt + \pgf@shorten@left}{\pgf@ya}}
\pgfpathlineto{\pgfpoint{\pgf@xb + 8pt - \pgf@shorten@right}{\pgf@ya}}
\pgfpathlineto{\pgfpoint{\pgf@xb + 8pt - \pgf@shorten@right}{\pgf@ya + 1.5 * \pgf@shorten@right}}
\pgfpathclose
}
}

\pgfdeclareshape{langpoint}{
\inheritsavedanchors[from=rectangle] 
\inheritanchorborder[from=rectangle]
\inheritanchor[from=rectangle]{center}
\inheritanchor[from=rectangle]{north}
\inheritanchor[from=rectangle]{south}
\inheritanchor[from=rectangle]{west}
\inheritanchor[from=rectangle]{east}
\backgroundpath{
\southwest \pgf@xa=\pgf@x \pgf@ya=\pgf@y
\northeast \pgf@xb=\pgf@x \pgf@yb=\pgf@y

\pgfmathsetmacro{\pgf@shorten@left}{\pgfkeysvalueof{/tikz/shorten left}}
\pgfmathsetmacro{\pgf@shorten@right}{\pgfkeysvalueof{/tikz/shorten right}}

\pgfpathmoveto{\pgfpoint{0.5 * (\pgf@xa + \pgf@xb)}{\pgf@ya - 2pt}}
\pgfpathlineto{\pgfpoint{\pgf@xa - 8pt}{\pgf@yb - 3 * \pgf@shorten@left + 5pt}} 
\pgfpathlineto{\pgfpoint{\pgf@xa - 8pt}{\pgf@yb -1pt}}
\pgfpathlineto{\pgfpoint{\pgf@xb + 8pt}{\pgf@yb -1pt}}
\pgfpathlineto{\pgfpoint{\pgf@xb + 8pt}{\pgf@yb - 3 * \pgf@shorten@left + 5pt}}
\pgfpathclose
}
}

\pgfdeclareshape{langcopoint}{
\inheritsavedanchors[from=rectangle] 
\inheritanchorborder[from=rectangle]
\inheritanchor[from=rectangle]{center}
\inheritanchor[from=rectangle]{north}
\inheritanchor[from=rectangle]{south}
\inheritanchor[from=rectangle]{west}
\inheritanchor[from=rectangle]{east}
\backgroundpath{
\southwest \pgf@xa=\pgf@x \pgf@ya=\pgf@y
\northeast \pgf@xb=\pgf@x \pgf@yb=\pgf@y

\pgfmathsetmacro{\pgf@shorten@left}{\pgfkeysvalueof{/tikz/shorten left}}
\pgfmathsetmacro{\pgf@shorten@right}{\pgfkeysvalueof{/tikz/shorten right}}

\pgfpathmoveto{\pgfpoint{0.5 * (\pgf@xa + \pgf@xb)}{\pgf@yb +0pt}}
\pgfpathlineto{\pgfpoint{\pgf@xa - 8pt}{\pgf@ya + 3 * \pgf@shorten@left - 5pt}} 
\pgfpathlineto{\pgfpoint{\pgf@xa - 8pt}{\pgf@ya + 1pt}}
\pgfpathlineto{\pgfpoint{\pgf@xb + 8pt}{\pgf@ya + 1pt}}
\pgfpathlineto{\pgfpoint{\pgf@xb + 8pt}{\pgf@ya + 3 * \pgf@shorten@left - 5pt}}
\pgfpathclose
}
}

\pgfdeclareshape{langrect}{
\inheritsavedanchors[from=rectangle] 
\inheritanchorborder[from=rectangle]
\inheritanchor[from=rectangle]{center}
\inheritanchor[from=rectangle]{north}
\inheritanchor[from=rectangle]{south}
\inheritanchor[from=rectangle]{west}
\inheritanchor[from=rectangle]{east}
\backgroundpath{
\southwest \pgf@xa=\pgf@x \pgf@ya=\pgf@y
\northeast \pgf@xb=\pgf@x \pgf@yb=\pgf@y

\pgfmathsetmacro{\pgf@shorten@left}{\pgfkeysvalueof{/tikz/shorten left}}
\pgfmathsetmacro{\pgf@shorten@right}{\pgfkeysvalueof{/tikz/shorten right}}

\pgfpathmoveto{\pgfpoint{\pgf@xa - 8pt}{\pgf@ya + 3 * \pgf@shorten@left - 5pt}} 
\pgfpathlineto{\pgfpoint{\pgf@xa - 8pt}{\pgf@ya + 1pt}}
\pgfpathlineto{\pgfpoint{\pgf@xb + 8pt}{\pgf@ya + 1pt}}
\pgfpathlineto{\pgfpoint{\pgf@xb + 8pt}{\pgf@ya + 3 * \pgf@shorten@left - 5pt}}
\pgfpathclose
}
}

\makeatother

\pgfkeyssetvalue{/tikz/shorten left}{0pt}
\pgfkeyssetvalue{/tikz/shorten right}{0pt}

\tikzstyle{kpoint common}=[draw,fill=white,inner sep=1pt,minimum height=4mm]

\tikzstyle{langstate}=[shape=langcopoint,shorten left=5pt,kpoint common,font=\footnotesize]
\tikzstyle{langeffect}=[shape=langpoint,shorten left=5pt,kpoint common,font=\footnotesize]
\tikzstyle{langstatedash}=[shape=langcopoint,dashed, shorten left=5pt,kpoint common,font=\footnotesize]
\tikzstyle{langeffectdash}=[shape=langpoint,dashed, shorten left=5pt,kpoint common,font=\footnotesize]
\tikzstyle{langbox}=[shape=langrect,shorten left=5pt,kpoint common,font=\footnotesize] 

\tikzstyle{kpoint}=[shape=cornerpoint,shorten left=5pt,kpoint common]
\tikzstyle{kpoint adjoint}=[shape=cornercopoint,shorten left=5pt,kpoint common]

\tikzstyle{kpoint conjugate}=[shape=cornerpoint,shorten right=5pt,kpoint common]
\tikzstyle{kpoint transpose}=[shape=cornercopoint,shorten right=5pt,kpoint common]
\tikzstyle{kpoint symm}=[shape=cornerpoint,shorten left=5pt,shorten right=5pt,kpoint common]

\tikzstyle{black kpoint}=[shape=cornerpoint,shorten left=5pt,kpoint common,fill=black,font=\color{white}]
\tikzstyle{black kpoint adjoint}=[shape=cornercopoint,shorten left=5pt,kpoint common,fill=black,font=\color{white}]
\tikzstyle{black kpointadj}=[shape=cornercopoint,shorten left=5pt,kpoint common,fill=black,font=\color{white}]

\tikzstyle{black dkpoint}=[shape=cornerpoint,shorten left=5pt,kpoint common,fill=black, doubled,font=\color{white}]
\tikzstyle{black dkpoint adjoint}=[shape=cornercopoint,shorten left=5pt,kpoint common,fill=black, doubled,font=\color{white}]
\tikzstyle{black dkpointadj}=[shape=cornercopoint,shorten left=5pt,kpoint common,fill=black, doubled,font=\color{white}]

\tikzstyle{kpointdag}=[kpoint adjoint]
\tikzstyle{kpointadj}=[kpoint adjoint]
\tikzstyle{kpointconj}=[kpoint conjugate]
\tikzstyle{kpointtrans}=[kpoint transpose]

\tikzstyle{big kpoint}=[kpoint, minimum width=1.2 cm, minimum height=8mm, inner sep=4pt, text depth=3mm]

\tikzstyle{wide kpoint}=[kpoint, minimum width=1 cm, inner sep=2pt]
\tikzstyle{wide kpointdag}=[kpointdag, minimum width=1 cm, inner sep=2pt]
\tikzstyle{wide kpointconj}=[kpointconj, minimum width=1 cm, inner sep=2pt]
\tikzstyle{wide kpointtrans}=[kpointtrans, minimum width=1 cm, inner sep=2pt]

\tikzstyle{gray kpoint}=[kpoint,fill=gray!50!white]
\tikzstyle{gray kpointdag}=[kpointdag,fill=gray!50!white]
\tikzstyle{gray kpointadj}=[kpointadj,fill=gray!50!white]
\tikzstyle{gray kpointconj}=[kpointconj,fill=gray!50!white]
\tikzstyle{gray kpointtrans}=[kpointtrans,fill=gray!50!white]

\tikzstyle{gray dkpoint}=[kpoint,fill=gray!50!white,doubled]
\tikzstyle{gray dkpointdag}=[kpointdag,fill=gray!50!white,doubled]
\tikzstyle{gray dkpointadj}=[kpointadj,fill=gray!50!white,doubled]
\tikzstyle{gray dkpointconj}=[kpointconj,fill=gray!50!white,doubled]
\tikzstyle{gray dkpointtrans}=[kpointtrans,fill=gray!50!white,doubled]

\tikzstyle{white label}=[draw,fill=white,rectangle,inner sep=0.7 mm]
\tikzstyle{gray label}=[draw,fill=gray!50!white,rectangle,inner sep=0.7 mm]
\tikzstyle{black label}=[draw,fill=black,rectangle,inner sep=0.7 mm]

\tikzstyle{dkpoint}=[kpoint,doubled]
\tikzstyle{wide dkpoint}=[wide kpoint,doubled]
\tikzstyle{dkpointdag}=[kpoint adjoint,doubled]
\tikzstyle{wide dkpointdag}=[wide kpointdag,doubled]
\tikzstyle{dkcopoint}=[kpoint adjoint,doubled]
\tikzstyle{dkpointadj}=[kpoint adjoint,doubled]
\tikzstyle{dkpointconj}=[kpoint conjugate,doubled]
\tikzstyle{dkpointtrans}=[kpoint transpose,doubled]

\tikzstyle{kscalar}=[kpoint common, shape=EBox, inner xsep=-1pt, inner ysep=3pt,font=\small]
\tikzstyle{kscalarconj}=[kpoint common, shape=WBox, inner xsep=-1pt, inner ysep=3pt,font=\small]

 \tikzstyle{upground}=[circuit ee IEC,ground,rotate=90,scale=2.5]
 \tikzstyle{downground}=[circuit ee IEC,ground,rotate=-90,scale=2.5]
 \tikzstyle{bigground}=[regular polygon,regular polygon sides=3,draw=gray,scale=0.50,inner sep=-0.5pt,minimum width=10mm,fill=gray]

\tikzstyle{arrs}=[-latex,font=\small,auto]
\tikzstyle{arrow plain}=[arrs]
\tikzstyle{arrow dashed}=[dashed,arrs]
\tikzstyle{arrow bold}=[very thick,arrs]
\tikzstyle{arrow hide}=[draw=white!0,-]
\tikzstyle{arrow reverse}=[latex-]
\tikzstyle{cdnode}=[]

%% file: figures/cap.tikz
\begin{tikzpicture}[scale=0.80]
	\begin{pgfonlayer}{nodelayer}
		\node [style=none] (0) at (1.25, -0.5) {};
		\node [style=none] (1) at (-1.25, -0.5) {};
	\end{pgfonlayer}
	\begin{pgfonlayer}{edgelayer}
		\draw [style=swap, bend left=90, looseness=1.50] (1.center) to (0.center);
	\end{pgfonlayer}
\end{tikzpicture} 

%% file: figures/cup.tikz
\begin{tikzpicture}[scale=0.80]
	\begin{pgfonlayer}{nodelayer}
		\node [style=none] (0) at (1.25, 0.5) {};
		\node [style=none] (1) at (-1.25, 0.5) {};
	\end{pgfonlayer}
	\begin{pgfonlayer}{edgelayer}
		\draw [style=swap, bend right=90, looseness=1.50] (1.center) to (0.center); 
	\end{pgfonlayer}
\end{tikzpicture}

%% file: figures/plaindot.tikz
\begin{tikzpicture}[scale=0.80]
	\begin{pgfonlayer}{nodelayer}
		\node [style=none] (0) at (0, -1) {};
		\node [style=none] (1) at (0, 1) {};
		\node [style=white dot] (2) at (0, 0) {};
		\node [style=none] (3) at (0, 0) {};
	\end{pgfonlayer}
	\begin{pgfonlayer}{edgelayer}
		\draw [style=swap] (0.center) to (1.center); 
	\end{pgfonlayer}
\end{tikzpicture}

%% file: figures/plain.tikz
\begin{tikzpicture}[scale=0.80]
	\begin{pgfonlayer}{nodelayer}
		\node [style=none] (0) at (0, -1) {};
		\node [style=none] (1) at (0, 1) {};
	\end{pgfonlayer}
	\begin{pgfonlayer}{edgelayer}
		\draw [style=swap] (0.center) to (1.center);  
	\end{pgfonlayer}
\end{tikzpicture}

%% file: figures/delete.tikz
\begin{tikzpicture}[scale=0.80]
	\begin{pgfonlayer}{nodelayer}
		\node [style=none] (0) at (0, 0.75) {};
		\node [style=white dot] (1) at (0, -0.5) {};
		\node [style=none] (2) at (0, -0.5) {};
	\end{pgfonlayer}
	\begin{pgfonlayer}{edgelayer}
		\draw [style=swap, in=90, out=-90, looseness=0.75] (0.center) to (1);   
	\end{pgfonlayer}
\end{tikzpicture}

%% file: figures/AND2.tikz
\begin{tikzpicture}[scale=0.80]
	\begin{pgfonlayer}{nodelayer}
		\node [style=none] (0) at (0, 0.5) {};
		\node [style=none] (1) at (0, -1.25) {};
		\node [style=langstate] (2) at (0, 0.75) {$Q\cap R$};
	\end{pgfonlayer}
	\begin{pgfonlayer}{edgelayer}
		\draw [style=swap] (0.center) to (1.center); 
	\end{pgfonlayer}
\end{tikzpicture}

%% file: figures/unknown.tikz
\begin{tikzpicture}[scale=0.80]
	\begin{pgfonlayer}{nodelayer}
		\node [style=none] (0) at (0, -0.75) {};
		\node [style=white dot] (1) at (0, 0.5) {};
		\node [style=none] (2) at (0, 0.5) {};
	\end{pgfonlayer}
	\begin{pgfonlayer}{edgelayer}
		\draw [style=swap, in=-90, out=90, looseness=0.75] (0.center) to (1); 
	\end{pgfonlayer}
\end{tikzpicture}

%% file: figures/AND3.tikz
\begin{tikzpicture}[scale=0.80]
	\begin{pgfonlayer}{nodelayer}
		\node [style=none] (0) at (0, 0.5) {};
		\node [style=none] (1) at (0, -1.25) {};
		\node [style=langstate] (2) at (0, 0.75) {$Q$};
	\end{pgfonlayer}
	\begin{pgfonlayer}{edgelayer}
		\draw [style=swap] (0.center) to (1.center);
	\end{pgfonlayer}
\end{tikzpicture}